\pdfoutput=1

\documentclass[11pt]{article}

\usepackage[]{ACL2023}

\usepackage{times}
\usepackage{latexsym}

\usepackage[T1]{fontenc}

\usepackage[utf8]{inputenc}

\usepackage{microtype}
\usepackage{multirow}
\usepackage{amsfonts}
\usepackage{graphicx}
\usepackage{amsmath}
\usepackage{subfigure}
\usepackage{amsmath}
\usepackage{inconsolata}
\usepackage{makecell}

\usepackage{pifont}

\title{Supervised Adversarial Contrastive Learning for Emotion Recognition \\ in Conversations}
\author{Dou Hu$^{1,2}$
        \and Yinan Bao$^{1,2}$
        \and Lingwei Wei$^{1,2}$
        \and Wei Zhou$^{1*}$
        \and Songlin Hu$^{1,2*}$
         \\
        $^{1}$ Institute of Information Engineering, Chinese Academy of Sciences \\
        $^{2}$ School of Cyber Security, University of Chinese Academy of Sciences  \\
        \texttt{\{hudou, baoyinan, weilingwei, zhouwei, husonglin\}@iie.ac.cn} \\
}

\begin{document}
\maketitle
\begingroup\def\thefootnote{*}\footnotetext{Corresponding author.}\endgroup
\begin{abstract}
Extracting generalized and robust representations is a major challenge in emotion recognition in conversations (ERC). To address this, we propose a supervised adversarial contrastive learning (SACL) framework for learning class-spread structured representations in a supervised manner. SACL applies contrast-aware adversarial training to generate worst-case samples and uses joint class-spread contrastive learning to extract structured representations. It can effectively utilize label-level feature consistency and retain fine-grained intra-class features. To avoid the negative impact of adversarial perturbations on context-dependent data, we design a contextual adversarial training (CAT) strategy to learn more diverse features from context and enhance the model's context robustness. Under the framework with CAT, we develop a sequence-based SACL-LSTM to learn label-consistent and context-robust features for ERC. Experiments on three datasets show that SACL-LSTM achieves state-of-the-art performance on ERC. Extended experiments prove the effectiveness of SACL and CAT.
\end{abstract}

\section{Introduction}
Emotion recognition in conversations (ERC) aims to detect emotions expressed by speakers during a conversation. 
The task is a crucial topic for developing empathetic machines \cite{DBLP:journals/inffus/MaNXC20}.
Existing works mainly focus on context modeling \cite{DBLP:conf/aaai/MajumderPHMGC19,DBLP:conf/emnlp/GhosalMPCG19,DBLP:conf/acl/HuWH20} and emotion representation learning \cite{DBLP:conf/acl/ZhuP0ZH20,DBLP:conf/aaai/YangSMC22,DBLP:conf/aaai/LiYQ22} to recognize emotions.
However, these methods have limitations in discovering the intrinsic structure of data relevant to emotion labels, and struggle to extract generalized and robust representations, resulting in mediocre recognition performance.

In the field of representation learning, label-based contrastive learning \cite{khosla2020supervised,lopez2022supervised} techniques are used to learn a generalized representation by capturing similarities between examples within a class and contrasting them with examples from other classes.
Since similar emotions often have similar context and overlapping feature spaces, these techniques that directly compress the feature space of each class are likely to hurt the fine-grained features of each emotion, thus limiting the ability of generalization.

To address these, 
we propose a supervised adversarial contrastive learning (SACL) framework to learn class-spread structured representations in a supervised manner.
SACL applies contrast-aware adversarial training to generate worst-case samples and uses a joint class-spread contrastive learning objective on both original and adversarial samples. It can effectively utilize label-level feature consistency and retain fine-grained intra-class features. 

Specifically, we adopt soft\footnote{The soft version means a cross-entropy term is added to alleviate the class collapse issue \cite{graf2021dissecting}, wherein each point in the same class has the same representation.} 
SCL \citep{gunel2020supervised} on original samples to obtain contrast-aware adversarial perturbations. Then, we put perturbations on the hidden layers to generate hard positive examples with a min-max training recipe. These generated samples can spread out the representation space for each class and confuse robust-less networks.
After that, we utilize a new soft
SCL on obtained adversarial samples to maximize the consistency of class-spread representations with the same label.
Under the joint objective on both original and adversarial samples, the network can effectively learn label-consistent features and achieve better generalization.

In context-dependent dialogue scenarios, directly generating adversarial samples interferes with the correlation between utterances, which is detrimental to context understanding. To avoid this, we design a contextual adversarial training (CAT) strategy to adaptively generate context-level worst-case samples and extract more diverse features from context. 
This strategy applies adversarial perturbations to the context-aware network structure in a multi-channel way, instead of directly putting perturbations on context-free layers in a single-channel way \cite{DBLP:journals/corr/GoodfellowSS14,DBLP:conf/iclr/MiyatoDG17}.
After introducing CAT, SACL can further learn more diverse features and smooth representation spaces from context-dependent inputs, as well as enhance the model's context robustness.

Under SACL framework, we design a sequence-based method SACL-LSTM to recognize emotion in the conversation. It consists of a dual long short-term memory (Dual-LSTM) module and an emotion classifier. 
Dual-LSTM is a modified version of the contextual perception module \cite{DBLP:conf/acl/HuWH20}, which can effectively capture contextual features from a dialogue.
With the guidance of SACL, the model can learn label-consistent and context-robust emotional features for the ERC task.

We conduct experiments on three public benchmark datasets. 
Results consistently demonstrate that our SACL-LSTM significantly outperforms other state-of-the-art methods on the ERC task, showing the effectiveness and superiority of our method. Moreover, extensive experiments prove that our SACL framework can capture better structured and robust representations for classification.

The main contributions are as follows:    
\textbf{1)} We propose a supervised adversarial contrastive learning (SACL) framework to extract class-spread structured representations for classification. It can effectively utilize label-level feature consistency and retain fine-grained intra-class features.
\textbf{2)} We design a contextual adversarial training (CAT) strategy to learn more diverse features from context-dependent inputs and enhancing the model’s context robustness.
\textbf{3)} We develop a sequence-based method SACL-LSTM under the framework to learn label-consistent and context-robust emotional features for ERC\footnote{To the best of our knowledge, this is the first attempt to introduce the idea of adversarial training into the ERC task.}. 
\textbf{4)} Experiments on three benchmark datasets show that SACL-LSTM significantly outperforms other state-of-the-art methods, and prove the effectiveness of the SACL framework\footnote{The source code is available at \url{https://github.com/zerohd4869/SACL}}.

\section{Methodology}
In this section, we first present the methodology of SACL framework.
Besides, for better adaptation to context-independent scenarios, we introduce a CAT strategy to SACL framework.
Finally, we apply the proposed SACL framework for emotion recognition in conversations and provide a sequence-based method SACL-LSTM.

\subsection{Supervised Adversarial Contrastive Learning Framework}
In the field of representation learning, label-based contrastive learning \cite{khosla2020supervised,lopez2022supervised} techniques are used to learn a generalized representation by capturing similarities between examples within a class and contrasting them with examples from other classes. However, directly compressing the feature space of each class is prone to harming fine-grained intra-class features, which limits the model’s ability to generalize. 

\begin{figure}[t]
  \centering
    \includegraphics[width=\linewidth]{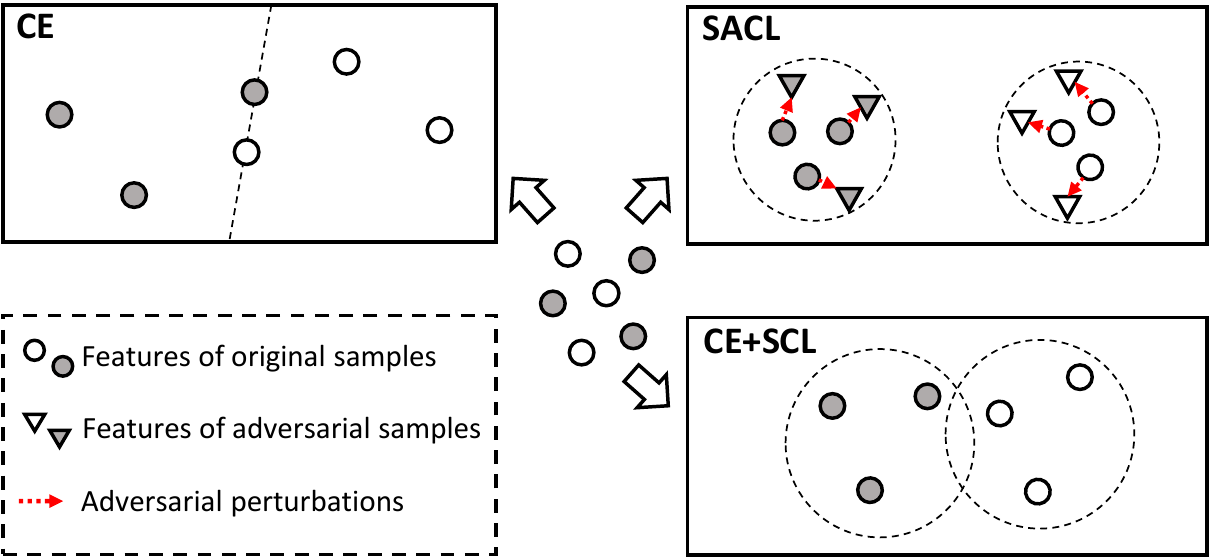}
    \caption{Comparison of different training objectives on a two-class case.
    CE and CE+SCL mean cross-entropy and soft supervised contrastive learning, respectively.
    }
  \label{fig:sacl_difference}
\end{figure}

To address this, we design a supervised adversarial contrastive learning (SACL) framework for learning class-spread structured representations. 
The framework applies contrast-aware adversarial training to generate worst-case samples and uses a joint class-spread contrastive learning objective on both original and adversarial samples. It can effectively utilize label-level feature consistency and retain fine-grained intra-class features. 
Figure~\ref{fig:sacl_difference} visualizes the difference between SACL and two representative optimization objectives (i.e., CE and soft SCL \citep{gunel2020supervised}) on a toy example.

\begin{figure*}[t]
  \centering
    \subfigure[Contextual Adversarial Training (CAT)]{
        \includegraphics[height=3.3cm]{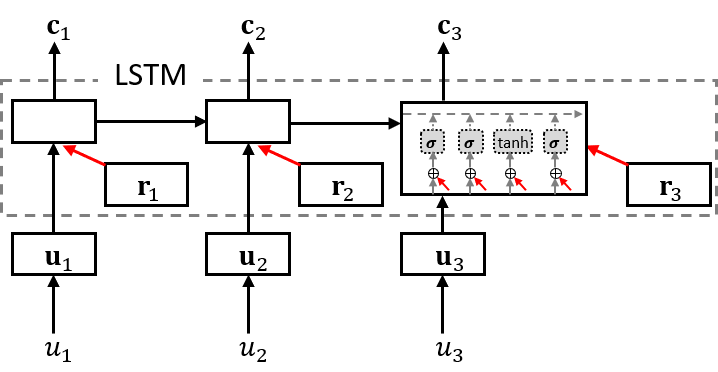}
        \label{fig:cat}
    } 
    \subfigure[Adversarial Training (AT)]{
        \includegraphics[height=3.3cm]{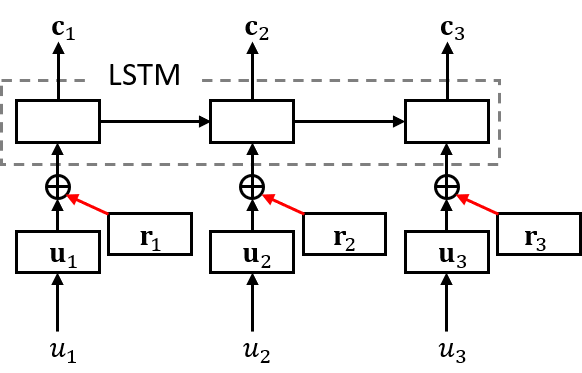}
        \label{fig:at}
    } 
    \subfigure[Vanilla Training (VT)]{
        \includegraphics[height=3.3cm]{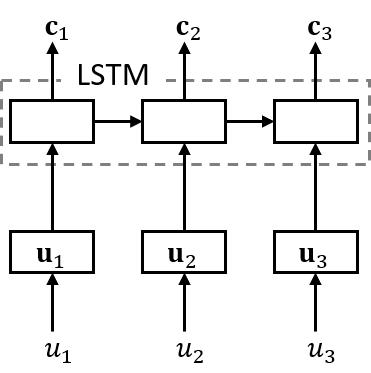}
        \label{fig:vt}
    } 
       \caption{
    An LSTM network against different perturbations.
    CAT, AT and VT represent the hidden layers with contextual adversarial perturbations $r_{\text{c-adv}}$, adversarial perturbations $r_{\text{adv}}$ and no perturbations, respectively.
    }
  \label{fig:perturb}
\end{figure*}

Formally, let us denote $I$ as the set of samples in a mini-batch.
Define $\phi(i) = \{ e \in I \backslash \{i\}: \hat{\mathbf{y}}_e =\hat{\mathbf{y}}_i \}$ is the set of indices of all positives in the mini-batch distinct from $i$, and $|\phi(i)|$ is its cardinality.
The loss function of soft SCL is a weighted average of CE loss and SCL loss with a trade-off scalar parameter $\lambda$, i.e., 
\begin{equation}
\mathcal{L}_{\text{soft-SCL}} = \mathcal{L}_{\text{CE}} + \lambda \mathcal{L}_{\text{SCL}},  \label{eq:softscl}
\end{equation}
where 
\begin{equation}
\mathcal{L}_{\text{CE}} = -   \sum\limits_{i \in I} {\mathbf{y}}_{i,k} \log (\hat{\mathbf{y}}_{i,k}), 
  \end{equation}   
\begin{equation}
\resizebox{0.89\linewidth}{!}{$
    \mathcal{L}_{\text{SCL}} = \sum\limits_{i \in I} \frac{-1}{|\phi(i)|}  \sum\limits_{e \in \phi(i)} \log \frac{\exp(sim(\mathbf{z}_i,\mathbf{z}_e) / \tau)  } {\sum\limits_{a\in A(i)} \exp(sim(\mathbf{z}_i, \mathbf{z}_a) / \tau) },
    $}
\end{equation}
$\mathbf{y}_{i,k}$ and $\hat{\mathbf{y}}_{i,k}$ denote the value of one-hot vector $\mathbf{y}_{i}$ and probability vector $\hat{\mathbf{y}}_{i}$ at class index k, respectively.
$A(i) = I \backslash \{i\} $. 
$\mathbf{z}_i$ refers to the hidden representation of the network's output for the $i$-th sample.
$sim(\cdot, \cdot)$ is a pairwise similarity function, 
i.e., dot product.
$\tau > 0$ is a scalar temperature parameter that controls the separation of classes.

At each step of training, we apply an adversarial training strategy with the soft SCL objective on original samples to produce anti-contrast worst-case samples. 
The training strategy can be implemented using a context-free approach such as FGM \citep{DBLP:conf/iclr/MiyatoDG17} or our context-aware CAT.  
These samples can be seen as hard positive examples, which spread out the representation space for each class and confuse the robust-less model.
After that, we utilize a new soft SCL on obtained adversarial samples to maximize the consistency of class-spread representations with the same label. 
Following the above calculation process 
of $\mathcal{L}_{\text{soft-SCL}}$ 
on original samples, the optimization objective on corresponding adversarial samples can be easily obtained in a similar way, i.e., $\mathcal{L}_{\text{soft-SCL}}^{\text{r-adv}}$.
 
The overall loss of SACL is defined as a sum of two soft SCL losses on both original and adversarial samples, i.e.,
\begin{equation}
     \mathcal{L} = \mathcal{L}_{\text{soft-SCL}} + \mathcal{L}_{\text{soft-SCL}}^{\text{r-adv}}. 
\end{equation}

\subsection{Contextual Adversarial Training}
Adversarial training (AT) \cite{DBLP:journals/corr/GoodfellowSS14,DBLP:conf/iclr/MiyatoDG17} is a widely used regularization method for models to improve robustness to small, approximately worst-case perturbations. In context-dependent scenarios, directly generating adversarial samples interferes with the correlation between samples, which is detrimental to context understanding.

To avoid this, we design a contextual adversarial training (CAT) strategy for a context-aware network, to obtain diverse context features and a robust model.
Different from the standard AT that put perturbations on context-free layers (e.g., word/sentence embeddings), we add adversarial perturbations to the context-aware network structure in a multi-channel way. Under a supervised training objective, it can obtain diverse features from context and enhance model robustness to contextual perturbations.

Let us denote $(u, y)$ as the mini-batch input sampled from distribution $D$ and $p(y|u; {\theta})$ as a context-aware model.
At each step of training, we identify the contextual adversarial perturbations $r_{\text{c-adv}}$ against the current model with the parameters $\hat{\theta}$, and put them on the context-aware hidden layers of the model.
With a linear approximation \cite{DBLP:journals/corr/GoodfellowSS14}, an $L_q$ norm-ball and a certain radius $\epsilon$ for $r_{\text{c-adv}}$, and a training objective $\ell$ (e.g., soft SCL), the formulation of CAT is illustrated by
\begin{equation}
\resizebox{0.89\linewidth}{!}{$
\begin{split}
& \min_{\theta}\mathbb{E}_{{(u,y)}\sim{D}}\max_{\| r_{\text{c-adv}}\|_q\leq\epsilon} \ell(u+r_{\text{c-adv}}, y; \theta),  \\
& \text{ where }
r_{\text{c-adv}} = - \epsilon {g} / {\| g \|_q},
\text{ } g = \nabla_u \log p(y \mid u; \hat{\theta}).   
\label{eq: AT}
\end{split}
$}
\end{equation}

Here, we take the LSTM network \citep{DBLP:journals/neco/HochreiterS97} with a sequence input $[u_1, u_2, ..., u_N]$ as an example, and the corresponding representations of the output are $[\mathbf{c}_1, \mathbf{c}_2, ..., \mathbf{c}_N]$.
Adversarial perturbations are put on context-aware hidden layers of the LSTM in a multi-channel way, including three gated layers and a memory cell layer in the LSTM structure, as shown in Figure~{\ref{fig:perturb}}.

With contextual perturbations on the network, there is a reasonable interpretation of the formulation in Eq.~\eqref{eq: AT}.
The inner maximization problem is finding the context-level worst-case samples for the network, and the outer minimization problem is to train a robust network to the worst-case samples.
After introducing CAT, our SACL can further learn more diverse features and smooth representation spaces from context-dependent inputs, as well as enhance the model's context robustness.

\subsection{Application for Emotion Recognition in Conversations}
\begin{figure}[t]
  \centering
    \includegraphics[width=\linewidth]{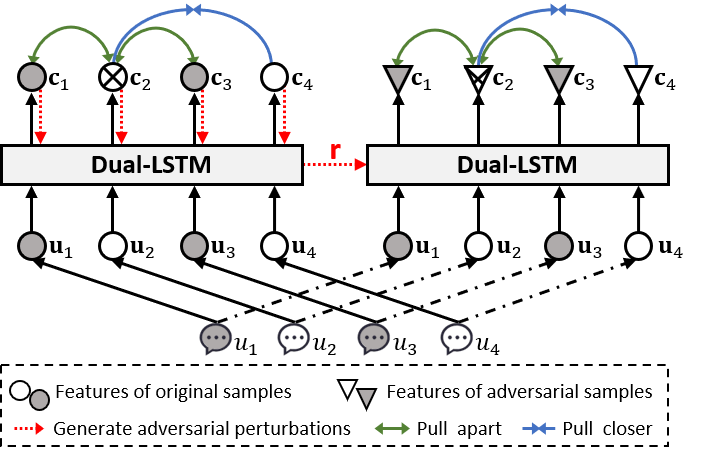}
    \caption{
    The overall architecture of SACL-LSTM. 
    The mark with/without shade means two different classes. We take the $\boldsymbol{\times}$-marked utterance as an example to show the objective of SACL.
    Dual-LSTM is a context-aware module that can capture sequential features in the conversation. 
    $r$ means contextual adversarial perturbations that put on hidden layers of Dual-LSTM.
    }
  \label{fig:model}
\end{figure}

In this subsection, we apply SACL framework to the task of emotion recognition in conversations (ERC), and present a sequence-based method SACL-LSTM. 
The overall architecture is illustrated in Figure~\ref{fig:model}.
With the guidance of SACL with CAT, the method can learn label-consistent and context-robust emotional features for better emotion recognition.

\subsubsection{Problem Statement}
The ERC task aims to recognize emotions expressed by speakers in a conversation. 
Formally, let $U=[u_1, u_2, ..., u_N]$ be a conversation with $N$ utterances and $M$ speakers/parties.
Each utterance $u_i$ is spoken by the party $p_{\phi(u_i)} \in \{p_1, p_2, ..., p_M\}$, where $\phi$ maps the utterance index into the corresponding speaker index.
For each $m \in [1,M]$, $U_{{m}}$ represents the set of utterances spoken by the party $p_{m}$, i.e., $U_{{m}} = \{ u_i | u_i \in U \text{ and }  u_i \text{ is spoken by }   p_{m}, \text{ } \forall i \in [1, N] \}$.
The goal is to identify the emotion label $y_i$ for each utterance $u_i$ from the pre-defined emotions $\mathcal{Y}$.

\subsubsection{Textual Feature Extraction}
Following previous works \cite{DBLP:conf/emnlp/GhosalMGMP20,DBLP:conf/acl/ShenWYQ20}, the pre-trained 
\textit{roberta-large}\footnote{\url{https://huggingface.co/}\label{huggingface}}
\cite{DBLP:journals/corr/abs-1907-11692} is fine-tuned on the train sets for utterance-level emotion classification, and then its parameters are frozen when training our model. 
Formally, given an utterance input $u_i$, the output of $[CLS]$ token in the last hidden layer of the encoder is used to obtain the utterance representation $\mathbf{u}_i$ with a dimension $d_u$.
We denote $\{ \mathbf{u}_i \in \mathbb{R}^{d_u} \}_{i=1}^N$ as the context-free textual features for $N$ utterances.

\subsubsection{Model Structure} \label{sec:modelst}

The network structure of SACL-LSTM consists of a dual long short-term memory (Dual-LSTM) module and an emotion classifier.

\paragraph{Dual-LSTM}
After extracting textual features, we design a Dual-LSTM module to capture situation- and speaker-aware contextual features in a conversation. It is a modified version of the contextual perception module in \citet{DBLP:conf/acl/HuWH20}.
 
Specifically, to alleviate the speaker cold-start issue\footnote{In multi-party interactions, some speakers have limited interaction with others, making it difficult to capture context-aware speaker characteristics directly with sequence-based networks, especially with the short speaker sequence.},
we modify the speaker perception module. 
If the number of utterances of the speaker is less than a predefined integer threshold $\xi$, the common characteristics of these cold-start speakers are directly represented by a shared general speaker vector $\mathbf{o}$.
The speaker-aware features $\mathbf{c}^{sp}_i$ are computed as:
\begin{equation}
\begin{split}
\resizebox{\linewidth}{!}{$\mathbf{c}^{sp}_i, \mathbf{h}_{m,j}^{sp} = \left\{
\begin{aligned}
&\mathbf{o}, \text{None} \ \ &\text{if} \ |U_{{m}}| < \xi, \\
&\overleftrightarrow{LSTM}^{sp} (\mathbf{u}_{i}, \mathbf{h}_{m,j-1}^{sp}),  j \in [1,|U_{m}|]&\text{otherwise},
\end{aligned}
\right.
$}
\end{split}
\end{equation} 
where $\overleftrightarrow{LSTM}^{sp}$ indicates a BiLSTM to obtain speaker embeddings. 
$\mathbf{h}^{sp}_{m, j}$ is the $j$-th hidden state of the party $p_m$ with a dimension of $d_h$. $m = \phi(u_i)$. 
$U_m$ refers to all utterances of $p_m$ in a conversation. 
The situation-aware features $\mathbf{c}^{si}_i$ are defined as,
\begin{equation}
    \mathbf{c}^{si}_i, \mathbf{h}^{si}_{i} = {\overleftrightarrow{LSTM}}^{si}(\mathbf{u}_{i}, \mathbf{h}^{si}_{i-1}), 
\end{equation}
where $\overleftrightarrow{LSTM}^{si}$ is a BiLSTM to obtain situation-aware embeddings and $\mathbf{h}^{si}_{i}$ is the hidden vector with a dimension of $d_h$.  

We concatenate the situation-aware and speaker-aware features to form the context representation of each utterance, i.e.,
$\mathbf{c}_i = [\mathbf{c}_i^{si};\mathbf{c}_{i}^{sp}].$

\paragraph{Emotion Classifier} \label{sec:emo}
Finally, according to the context representation, an emotion classifier is applied to predict the emotion label of each utterance. 
\begin{equation}
    \hat{\mathbf{y}}_i = softmax(\mathbf{W}_c \mathbf{c}_i + \mathbf{b}_c),    
\end{equation} 
where $\mathbf{W}_c \in \mathbb{R}^{4d_h \times |\mathcal{Y}|}$ and $\mathbf{b}_c \in \mathbb{R}^{|\mathcal{Y}|}$ 
are trainable parameters.
$|\mathcal{Y}|$ is the number of emotion labels.

\subsubsection{Optimization Process} 
Under SACL framework, we apply contrast-aware CAT to generate worst-case samples and utilize a joint class-spread contrastive learning objective on both original and adversarial samples. 
At each step of training, we apply the CAT strategy with the soft SCL objective on original samples to produce context-level adversarial perturbations.
The perturbations are put on context-aware hidden layers of Dual-LSTM in a multi-channel way, and then obtain adversarial samples.
After that, we leverage a new soft SCL on these worst-case  samples to maximize the consistency of emotion-spread representations with the same label.
Under the joint objective on both original and adversarial samples, SACL-LSTM can learn label-consistent and context-robust emotional features for ERC.

\section{Experimental Setups}
\subsection{Datasets}
We evaluate our model on three benchmark datasets.
\textbf{IEMOCAP} \cite{DBLP:journals/lre/BussoBLKMKCLN08} contains dyadic conversation videos between pairs of ten unique speakers, where the first eight speakers belong to train sets and the last two belong to test sets.
The utterances are annotated with one of six emotions, namely happy, sad, neutral, angry, excited, and frustrated. 
\textbf{MELD} \cite{DBLP:conf/acl/PoriaHMNCM19} contains multi-party conversation videos collected from Friends TV series. Each utterance is annotated with one of seven emotions, i.e., joy, anger, fear, disgust, sadness, surprise, and neutral.
\textbf{EmoryNLP} \cite{DBLP:conf/aaai/ZahiriC18} is a textual corpus that comprises multi-party dialogue transcripts of the Friends TV show. Each utterance is annotated with one of seven emotions, 
i.e., sad, mad, scared, powerful, peaceful, joyful, and neutral.

The statistics are reported in Table \ref{tab:datasets}. In this paper, we focus on ERC in a textual setting. Other multimodal knowledge (i.e., acoustic and visual modalities) is not used.
We use the pre-defined train/val/test splits in MELD and EmoryNLP.
Following previous studies \cite{DBLP:conf/naacl/HazarikaPZCMZ18,DBLP:conf/emnlp/GhosalMPCG19}, we randomly extract 10\% of the training dialogues in IEMOCAP as validation sets since there is no predefined train/val split.

\begin{table}[t]
\centering  
\resizebox{\linewidth}{!}{$
\begin{tabular}{lcccccccc}
  \hline
  \multirow{2}{*}{\textbf{Dataset}} &  \multicolumn{3}{ c }{\textbf{\# Dialogues}}  & \multicolumn{3}{c}{\textbf{\# Utterances}} & 
  \multirow{1}{*}{\textbf{\# Avg.} } & 
  \multirow{1}{*}{\textbf{\# Avg.} }  
  \\
    &  \textbf{{train}} & \textbf{{val}} & \textbf{{test}}
    &  \textbf{{train}} & \textbf{{val}}   & \textbf{{test}} 
    &  \multirow{1}{*}{\textbf{Turns} } &  \multirow{1}{*}{\textbf{Parties} }  
    \\
  \hline
  IEMOCAP & \multicolumn{2}{c}{120} & 31 & \multicolumn{2}{c}{5810} & 1623 & 49.2 & 2 
  \\  
  MELD & 1039 & 114 & 280 & 9989 &  1109 & 2610 & 9.6 & 2.7 
  \\ 
  EmoryNLP & 659 & 89 & 79 &  7551 & 954 & 984  & 11.5   &  3.2   
  \\ \hline
  \end{tabular}
  $}
  \caption{The statistics of three datasets. }
  \label{tab:datasets}
\end{table}

\subsection{Comparison Methods}
The fourteen baselines compared are as follows.
1) Sequence-based methods:
\textbf{bc-LSTM} \cite{DBLP:conf/acl/PoriaCHMZM17}
employs an utterance-level LSTM to capture contextual features.
\textbf{DialogueRNN} \cite{DBLP:conf/aaai/MajumderPHMGC19} 
is a recurrent network to track speaker states and context.
\textbf{COSMIC} \cite{DBLP:conf/emnlp/GhosalMGMP20} 
uses GRUs to incorporate commonsense knowledge and capture complex interactions.
\textbf{DialogueCRN} \cite{DBLP:conf/acl/HuWH20} 
is a cognitive-inspired network with multi-turn reasoning modules that captures implicit emotional clues in a dialogue.
\textbf{CauAIN}  \cite{DBLP:conf/ijcai/ZhaoZL22}
uses causal clues in commonsense knowledge to enrich the modeling of speaker dependencies.

2) Graph-based methods:
\textbf{DialogueGCN} \cite{DBLP:conf/emnlp/GhosalMPCG19} 
uses GRUs and GCNs with relational edges to capture context and speaker dependency.
\textbf{RGAT} \cite{DBLP:conf/emnlp/IshiwatariYMG20} 
applies position encodings to RGAT to consider speaker and sequential dependency. 
\textbf{DAG-ERC} \cite{DBLP:conf/acl/ShenWYQ20} 
adopts a directed GNN to model the conversation structure.
\textbf{SGED+DAG} \cite{DBLP:conf/ijcai/BaoMWZH22} 
is a speaker-guided framework with a one-layer DAG that can explore complex speaker interactions.

3) Transformer-based methods:
\textbf{KET} \cite{DBLP:conf/emnlp/ZhongWM19} 
incorporates commonsense knowledge and context into a Transformer.
\textbf{DialogXL} \cite{DBLP:conf/aaai/ShenCQX21} 
adopts a modified XLNet to deal with longer context and multi-party structures.
\textbf{TODKAT} \cite{DBLP:conf/acl/ZhuP0ZH20} enhances the ability of Transformer by incorporating commonsense knowledge and a topic detection task.
\textbf{CoG-BART} \cite{DBLP:conf/aaai/LiYQ22}
uses a SupCon loss \cite{khosla2020supervised} and a response generation task to enhance BART's ability.
\textbf{SPCL+CL} \cite{DBLP:conf/emnlp/SongXH22}
applies a prompt-based BERT with supervised prototypical contrastive learning \cite{DBLP:conf/cvpr/00230W0W21,lopez2022supervised} and curriculum learning \cite{DBLP:conf/icml/BengioLCW09}.

\begin{table*}[t]
    \centering
    \resizebox{0.96\linewidth}{!}{$
    \begin{tabular}{l|r|cc|cc|cc|cc}
    \hline
     \multicolumn{1}{c|}{\multirow{2}{*}{Methods}}
      & \multicolumn{1}{c|}{\multirow{2}{*}{\# Param.}}
      & \multicolumn{2}{c|}{IEMOCAP}
      & \multicolumn{2}{c|}{MELD}
      & \multicolumn{2}{c|}{EmoryNLP}
      & \multicolumn{2}{c}{Avg.}
      \\ 
      &
      &  Acc &    w-F1 
      &  Acc &    w-F1
      &  Acc &    w-F1 
      &  Acc &    w-F1  \\ 
      \hline
    \multicolumn{1}{l}{{\it Transformer-based Methods}}  \\ 
      KET$^\dag$$^\ddag$  \cite{DBLP:conf/emnlp/ZhongWM19}
      & -
      & - & 59.56 & - & 58.18 & - & 34.39  & - & 50.17 \\ 
       DialogXL$^\ddag$ \cite{DBLP:conf/aaai/ShenCQX21}
      & -
      &  - & 65.94 &  - & 62.41 & -  & 34.73 & - & 54.36 \\ 
      TODKAT$^\dag$$^\ddag$  \cite{DBLP:conf/acl/ZhuP0ZH20}
      & - 
      & 61.11 &61.33 	&67.24 		&65.47 	&42.38 		&38.69 	&56.91 	&55.16 \\ 
      CoG-BART \cite{DBLP:conf/aaai/LiYQ22}
      & 415.1M
      & 65.02 & 64.87 & 64.95 & 63.82 & 40.94 & 37.33  &56.97 & 55.34 \\ 
      SPCL+CL \cite{DBLP:conf/emnlp/SongXH22}
      & 356.7M
      & 66.71 & 66.93 & 64.36 & 64.93 & 40.32 & 39.45 &57.13 & 57.10 \\  
      \hline
     \multicolumn{1}{l}{{\it Graph-based Methods}}  \\ 
      DialogueGCN  \cite{DBLP:conf/emnlp/GhosalMPCG19} 
      & 2.1M
      & 62.49 & 62.11 & 63.62 & 62.68 & 36.87 & 34.63 &54.33 & 53.14 \\ 
      RGAT$^\ddag$ \cite{DBLP:conf/emnlp/IshiwatariYMG20}
      & -
      & - & 65.22 & - & 60.91 & - & 34.42  & - & 53.52 \\ 
      DAG-ERC \cite{DBLP:conf/acl/ShenWYQ20}
      & 9.5M
      & 66.54 & 66.53 & 63.75 & 63.36 & 39.64 & 38.29  &56.64 & 56.06 \\
      SGED+DAG \cite{DBLP:conf/ijcai/BaoMWZH22}
      & 3.0M
      & 66.29 & 66.27 & 63.60 & 63.16 & 39.19 & 38.11 &56.36 & 55.85 \\ 
      \hline 
      \multicolumn{1}{l}{{\it Sequence-based Methods}}  \\ 
      bc-LSTM \cite{DBLP:conf/acl/PoriaCHMZM17}
      & 1.2M
      & 63.08 & 62.84 & 65.87 & 64.87 & 40.85 & 36.84 &56.60 & 54.85 \\ 
      DialogueRNN \cite{DBLP:conf/aaai/MajumderPHMGC19}
      & 9.9M
      & 64.85 & 64.65 & 65.96 & 65.30 & \textbf{43.66} & 37.54 &58.16 & 55.83 \\    
       COSMIC$^\dag$ \cite{DBLP:conf/emnlp/GhosalMGMP20}
      & 11.9M
      & 63.43 & 63.43 & 65.96 & 65.03 & 41.79 & 38.49 &57.06 & 55.65 \\ 
      DialogueCRN \cite{DBLP:conf/acl/HuWH20} 
      & 3.3M
      & 67.39 & 67.53 & 66.93 & 65.77 & 41.04 & 38.79  &58.45 & 57.36 \\  
     CauAIN$^\dag$  \cite{DBLP:conf/ijcai/ZhaoZL22}
      & 6.1M
      & 65.08 & 65.01 & 65.85 & 64.89 & 43.13 & 37.87 &58.02 & 55.92 \\  
      \textbf{SACL-LSTM} (ours)            
      & 2.6M
      & \textbf{69.08}$^*$ & \textbf{69.22}$^*$ & \textbf{67.51}$^*$ & \textbf{66.45}$^*$ & 42.21 & \textbf{39.65}$^*$ & \textbf{59.60}$^*$ & \textbf{58.44}$^*$ \\
      \hline
    \end{tabular}
    $}
    \caption{Overall results
    (\%) against various methods for ERC. 
    We present accuracy (Acc) and weighted-F1 (w-F1) score for each dataset.
    $\dag$ means the external knowledge is used.
    \# Param. means the average number of learnable model parameters.
    $\ddag$ means the results are from the original paper or their official repository; results of CoG-BART and SPCL+CL are reproduced under model initialization with 
    \textit{bart-large}\textsuperscript{\ref{huggingface}}
    and  \textit{roberta-large}\textsuperscript{\ref{huggingface}},
    respectively; all other results are reproduced using \textit{roberta-large} features that our SACL-LSTM uses. 
    For each reproduced method, we run five random seeds and report the average result on test sets. Best results are highlighted in bold. {*} represents statistical significance over state-of-the-art scores under the paired t-test (p<0.05).
    }
    \label{tab:result}
\end{table*}

\subsection{Evaluation Metrics}
Following previous works \cite{DBLP:conf/acl/HuWH20,DBLP:conf/aaai/LiYQ22}, we report the accuracy and weighted-F1 score to measure the overall performance. 
Also, the F1 score per class and macro-F1 score are reported to evaluate the fine-grained performance.
For the structured representation evaluation, we choose three supervised clustering metrics (i.e., ARI, NMI, and FMI) and three unsupervised clustering metrics (i.e., SC, CHI, and DBI) to measure the clustering performance of learned representations.
For the empirical robust evaluation \cite{DBLP:conf/sp/Carlini017}, we use the robust weighted-F1 score on adversarial samples generated from original test sets.
Besides, the paired t-test \cite{kim2015t} is used to verify the statistical significance of the differences between the two approaches.

\subsection{Implementation Details}
All experiments are conducted on a single NVIDIA Tesla V100 32GB card. 
The validation sets are used to tune hyperparameters and choose the optimal model.
For each method, we run five random seeds and report the average result of the test sets.
The network parameters of our model are optimized by using Adam optimizer \citep{DBLP:journals/corr/KingmaB14}.
More experimental details are listed in Appendix~\ref{sec:appendix:setups}.

\section{Results and Analysis} \label{sec:sec:exp}
\begin{table}[t]
\centering
\subtable[{IEMOCAP}]{
\resizebox{\linewidth}{!}{$
\begin{tabular}{l|cccccc|c}
\hline     
\multicolumn{1}{c|}{\multirow{1}{*}{Methods}} &  {\it Hap.} & {\it Sad.} & {\it Neu.} & {\it Ang.} & {\it Exc.} & {\it Fru.}  & Avg.  \\
\hline
DialogueCRN & 54.28 & 81.34 & 69.57 & 62.09 & 67.33 & 64.22 & 66.47 \\ 
\textbf{SACL-LSTM} & \textbf{56.91}$^*$ & \textbf{84.78}$^*$ & \textbf{70.00}$^*$ & \textbf{64.09}$^*$ & \textbf{69.70}$^*$ & \textbf{65.02}$^*$  & \textbf{68.42}$^*$ \\  
Improve & +2.63	&+3.44	&+0.43	&+2.00   &+2.37    &+0.80    &+1.95  \\ \hline
\end{tabular}
$}
\label{tab:subtab11}
}
\subtable[MELD]{
\resizebox{\linewidth}{!}{$
\begin{tabular}{l|ccccccc|c}
\hline
\multicolumn{1}{c|}{\multirow{1}{*}{Methods}} &  {\it Neu.} & {\it Sur.} & {\it Fea.} & {\it Sad.} & {\it Joy.} & {\it Dis.} & {\it Ang.} & Avg.  \\  \hline
DialogueCRN & 79.72	&57.62	&18.26 	&39.30	&64.56 &\textbf{32.07}	&\textbf{52.53}	&49.15  \\ 
\textbf{SACL-LSTM} &\textbf{80.17}$^*$	&\textbf{58.77}$^*$	&\textbf{26.23}$^*$ 	&\textbf{41.34}$^*$	&\textbf{64.98}$^*$	&31.47	&52.35	&\textbf{50.76}$^*$ \\ 
Improve &+0.45	&+1.15	&+7.97 	&+2.04	&+0.42	&-0.60	&-0.18	&+1.61 \\ \hline
\end{tabular}
$}
\label{tab:subtab12}
}
\subtable[EmoryNLP]{
\resizebox{\linewidth}{!}{$
\begin{tabular}{l|ccccccc|c}
\hline
\multicolumn{1}{c|}{\multirow{1}{*}{Methods}} & {\it Joy.} & {\it Mad.} & {\it Pea.} & {\it Neu.} & {\it Sad.} & {\it Pow.} & {\it Sca.} & Avg.  \\
\hline
DialogueCRN & 54.42	&36.44	&10.18	&53.83	&25.74	&4.55	&\textbf{37.49}  &31.81 \\ 
\textbf{SACL-LSTM} &\textbf{54.78}$^*$	&\textbf{37.68}$^*$	&\textbf{11.66}$^*$	&\textbf{55.42}$^*$	&\textbf{25.83}	&\textbf{5.43}$^*$	&37.11	&\textbf{32.56}$^*$ \\ 
Improve &+0.36	&+1.24	&+1.48	&+1.59	&+0.09	&+0.88	&-0.38	&+0.75 \\ \hline
\end{tabular}
$}
\label{tab:subtab13}
}
\caption{Fine-grained results (\%) of SACL-LSTM and DialogueCRN for all emotion categories. DialogueCRN is the sub-optimal method in Table~\ref{tab:result}. We report F1 score per class and macro-F1 score.}
\label{tab:fine-grained_result}
\end{table}

\subsection{Overall Results}
The overall results\footnote{We noticed that DialogueRNN and CauAIN present a poor weighted-F1 but a fine accuracy score on EmoryNLP, which is most likely due to the highly class imbalance issue.} are reported in Table~\ref{tab:result}. 
SACL-LSTM consistently obtains the best weighted-F1 score over comparison methods on three datasets.
Specifically, SACL-LSTM obtains \textbf{+1.1\%} absolute improvements over other state-of-the-art methods in terms of the average weighted-F1 score on three datasets. 
Besides, SACL-LSTM obtains \textbf{+1.2\%} absolute improvements in terms of the average accuracy score.
The results indicates the good generalization ability of our method to unseen test sets.

We also report fine-grained results on three datasets in Table~\ref{tab:fine-grained_result}. 
SACL-LSTM achieves better results for most emotion categories (17 out of 20 classes), except three classes (i.e., disgust and anger in MELD, and scared in EmoryNLP). 
It is worth noting that SACL-LSTM obtains \textbf{+2.0\%}, \textbf{+1.6\%} and \textbf{+0.8\%} absolute improvements in terms of the macro-F1 (average score of F1 for all classes) on IEMOCAP, MELD and EmoryNLP, respectively.

\subsection{Ablation Study}
We conduct ablation studies to evaluate key components in SACL-LSTM. 
The results are shown in Table~\ref{tab:abla}.
When removing the proposed SACL framework (i.e., {- w/o SACL}) 
and replacing it with a simple cross-entropy objective, we obtain inferior performance in terms of all metrics. 
When further removing the context-aware Dual-LSTM module (i.e., {- w/o SACL - w/o Dual-LSTM}) and replacing it with a context-free MLP (i.e., a fully-connected neural network with a single hidden layer), the results decline significantly on three datasets. 
It shows the effectiveness of both components.

\begin{table}[t]
    \centering
    \resizebox{\linewidth}{!}{$
    \begin{tabular}{l|c|c|c}
    \hline
       \multicolumn{1}{c|}{\multirow{1}{*}{Methods}}
      & \multicolumn{1}{c|}{IEMOCAP}
      & \multicolumn{1}{c|}{MELD}
      & \multicolumn{1}{c}{EmoryNLP}
     \\ \hline
      \textbf{SACL-LSTM}                 
      & \textbf{69.22}\scriptsize{$\pm$0.54} 
      & \textbf{66.45}\scriptsize{$\pm$0.35} 
      & \textbf{39.65}\scriptsize{$\pm$0.66}  \\
      \ \ - w/o SACL                    
      & 68.17\scriptsize{$\pm$0.63} 
      & 65.64\scriptsize{$\pm$0.14} 
      & 38.65\scriptsize{$\pm$0.62}  \\ 
      \ \ - w/o SACL - w/o Dual-LSTM          
      & 52.99\scriptsize{$\pm$0.49} 
      & 64.65\scriptsize{$\pm$0.12} 
      & 37.74\scriptsize{$\pm$0.20} \\ 
      \hline
    \end{tabular}
    $}
    \caption{
    Ablation results (\%) of SACL-LSTM.
    ``- w/o SACL'' means replacing the SACL with a cross-entropy term.
    ``- w/o Dual-LSTM'' means replacing the Dual-LSTM with an MLP.
    We report the average score and standard deviation of the weighted-F1 with five seeds.
    }
    \label{tab:abla}
\end{table}

\subsection{Comparison with Different Optimization Objectives}
To demonstrate the superiority of SACL, we include control experiments that replace it with the following optimization objectives, i.e., CE+SCL (soft SCL) \citep{gunel2020supervised}, CE+SupCon\footnote{The idea of SupCon is very similar to SCL. Their implementations are slightly different. Combined with CE, they achieved very close performance, as shown in Table~\ref{tab:cl_model}.}
\cite{khosla2020supervised}, and cross-entropy (CE). 

Table~\ref{tab:cl_model} shows results against various optimization objectives.
SACL significantly outperforms the comparison objectives on three datasets. 
CE+SCL and CE+SupCon objectives apply label-based contrastive learning to extract a generalized representation, leading to better performance than CE.
However, they compress the feature space of each class and harm fine-grained intra-class features, yielding inferior results than our SACL.
SACL uses a joint class-spread contrastive learning objective on both original and adversarial samples. It can effectively utilize label-level feature consistency and retain fine-grained intra-class features.

\begin{table}[t]
\centering
    \resizebox{0.83\linewidth}{!}{$
  \begin{tabular}{l|ccc}
    \hline
  \multicolumn{1}{c|}{\multirow{2}{*}{\makecell[c]{Optimization \\ Objectives}} } & \multicolumn{1}{c}{\multirow{2}{*}{IEMOCAP}} & \multicolumn{1}{c}{\multirow{2}{*}{MELD}} & \multicolumn{1}{c}{\multirow{2}{*}{EmoryNLP}}  \\  
  & \\
\hline
 \textbf{SACL}                 
      & \textbf{69.22}\scriptsize{$\pm$0.54} 
      & \textbf{66.45}\scriptsize{$\pm$0.35} 
      & \textbf{39.65}\scriptsize{$\pm$0.66}  \\
CE+SCL
& 68.32\scriptsize{$\pm$0.45} &65.95\scriptsize{$\pm$0.20} &38.93\scriptsize{$\pm$0.89} \\
CE+SupCon
& 68.37\scriptsize{$\pm$0.36} &65.89\scriptsize{$\pm$0.38} &39.00\scriptsize{$\pm$0.93} \\
CE & 68.17\scriptsize{$\pm$0.63} &65.64\scriptsize{$\pm$0.14} &38.65\scriptsize{$\pm$0.62} \\
\hline
\end{tabular}
$}
\caption{Comparison results (\%) against different optimization objectives. 
    We report the weighted-F1 score.
  }
 \label{tab:cl_model}
\end{table}

\begin{table}[t]
\centering
    \resizebox{0.8\linewidth}{!}{$
  \begin{tabular}{l|ccc}
    \hline
  \multicolumn{1}{c|}{\multirow{2}{*}{\makecell[c]{Training \\ Strategies}} } & \multicolumn{1}{c}{\multirow{2}{*}{IEMOCAP}} & \multicolumn{1}{c}{\multirow{2}{*}{MELD}} & \multicolumn{1}{c}{\multirow{2}{*}{EmoryNLP}}  \\  
  \\
\hline
\textbf{SACL} \\
\ \ - w/ \textbf{CAT}                
& \textbf{69.22}\scriptsize{$\pm$0.54} 
& \textbf{66.45}\scriptsize{$\pm$0.35} 
& \textbf{39.65}\scriptsize{$\pm$0.66}  \\
\ \ - w/ CRT 
& 68.28\scriptsize{$\pm$0.72} & 65.70\scriptsize{$\pm$0.29} & 39.16\scriptsize{$\pm$0.59} \\
\ \ - w/ AT
& 68.95\scriptsize{$\pm$1.03} & 65.69\scriptsize{$\pm$0.34} & 38.58\scriptsize{$\pm$0.39} \\
\ \ - w/ VT 
& 68.32\scriptsize{$\pm$0.45} & 65.95\scriptsize{$\pm$0.20} & 38.93\scriptsize{$\pm$0.89} \\
\hline
\end{tabular}
$}
\caption{Comparison results (\%) against different training strategies under the SACL framework.
CAT, CRT, AT, and VT are contextual adversarial training, contextual random training, adversarial training, and vanilla training, respectively.
We report the weighted-F1 score.
}
  \label{tab:perturbations}
\end{table}

\subsection{Comparison with Different Training Strategies} 
To evaluate the effectiveness of contextual adversarial training (CAT), 
we compare with different training strategies, i.e., adversarial training (AT) \cite{DBLP:conf/iclr/MiyatoDG17}, contextual random training (CRT), and vanilla training (VT). CRT is the strategy in which we replace $r_{\text{c-adv}}$ in CAT with random perturbations from a multivariate Gaussian with the scaled norm on context-aware hidden layers.

The results are reported in Table~\ref{tab:perturbations}. 
Compared with other strategies, our CAT obtains better performance consistently on three datasets.
It shows that CAT can enhance the diversity of emotional features by adding adversarial perturbations to the context-aware structure with a min-max training recipe.
We notice that AT strategy achieves the worst performance on MELD and EmoryNLP with the extremely short length of conversations. It indicates that AT is difficult to improve the diversity of context-dependent features with a limited context.

\begin{table}[t]
\centering
\resizebox{1\linewidth}{!}{$
\begin{tabular}{c|l|ccc|ccc}
\hline
 \multicolumn{1}{c|}{\multirow{3}{*}{\rotatebox{90}{\small{ }}}} & \multicolumn{1}{c|}{\multirow{2}{*}{\makecell[c]{Optimization \\
Objectives}}} & 
 \multicolumn{3}{c|}{Supervised Metrics} & \multicolumn{3}{c}{Unsupervised Metrics} 
 \\ 
 & & ARI$\uparrow$ & NMI$\uparrow$  & FMI$\uparrow$ & SC$\uparrow$ & \multicolumn{1}{c}{CHI$\uparrow$ } & DBI$\downarrow$  \\ 
 &  & (\%) & (\%) & (\%) &  \\ \hline
\multirow{4}{*}{\rotatebox{90}{\small{{{IEMOCAP}}}}}
    & CE   & 40.61 & 47.39 & 51.56 & 0.36  & 1231.19  & 1.06   \\ 
    & CE+SCL & 40.55 & 47.25 & 51.53 & 0.36  & 1188.46  & 1.07  \\ 
    & \textbf{SACL}            
    & \textbf{41.95} & \textbf{48.26} & \textbf{52.62} & \textbf{0.39}  & \textbf{1696.05}  & \textbf{0.99}   \\ 
    & Improve & +1.34 & +0.87 & +1.06 & +0.03  & +464.86  & +0.07  \\ 
    \hline
\multicolumn{1}{l|}{\multirow{4}{*}{\rotatebox{90}{\small{{MELD}}}}} 
    & CE    & 40.74 & 27.00 & 59.41 & 0.24  & 755.83  & 1.41  \\ 
    & CE+SCL & 40.92 & 27.13 & 59.52 & 0.23  & 738.70 & 1.42  \\
    & \textbf{SACL}        & \textbf{42.34} & \textbf{28.22} & \textbf{60.42} & \textbf{0.31}  & \textbf{1342.38}  & \textbf{1.16}  \\
    & Improve & +1.42 & +1.09 & +0.90 & +0.07  & +586.55  & +0.25  \\ 
    \hline
\end{tabular}
$}
\caption{Clustering results against different optimization objectives. 
Adjusted Rand Index (ARI), Normalized Mutual Information (NMI),  and Fowlkes-Mallows Index (FMI) evaluate the accuracy of clustering. Silhouette Coefficient (SC), Calinski-Harabasz Index (CHI), and Davies-Bouldin Index (DBI) evaluate the separation and compactness of clustering.
SC, CHI, and DBI are evaluated based on K-Means, and we define the number of clusters $K$ as the true number of categories, i.e., $K=6$ for IEMOCAP, and $K=7$ for MELD.}
\label{tab:clustering-performance}
\end{table}

\subsection{Structured Representation Evaluation}
To evaluate the quality of structured representations, we measure the clustering performance based on the representations learned with different optimization objectives on the test set of IEMOCAP and MELD.
Table~\ref{tab:clustering-performance} reports the clustering results of the Dual-LSTM network under three optimization objectives, including CE,  CE+SCL, and our SACL.

According to supervised clustering metrics, the proposed SACL outperforms other optimization objectives by \textbf{+1.3\%} and \textbf{+1.4\%} in ARI, \textbf{+0.9\%} and \textbf{+1.1\%} in NMI, \textbf{+1.1\%} and \textbf{+0.9\%} in FMI for  IEMOCAP and MELD, respectively.
The more accurate clustering results show that our SACL can distinguish different data categories and assign similar data points to the same categories. It indicates that SACL can discover the intrinsic structure of data relevant to labels and extract generalized representations for emotion recognition.

According to unsupervised clustering metrics, SACL achieves better results than other optimization objectives by \textbf{+0.03} and \textbf{+0.07} in SC, \textbf{+464.86} and \textbf{+586.55} in CHI, and \textbf{+0.07} and \textbf{+0.25} in DBI for IEMOCAP and MELD, respectively.
Better performance on these metrics suggests that SACL can learn more clear, separated, and compact clusters.
This indicates that SACL can better capture the underlying structure of the data, which can be beneficial for subsequent emotion recognition.

Overall, the results demonstrate the effectiveness of the SACL framework in learning structured representations for improving clustering performance and quality, as evidenced by the significant improvements in various clustering metrics.

\begin{figure}[t]
  \centering
    \includegraphics[width=\linewidth]{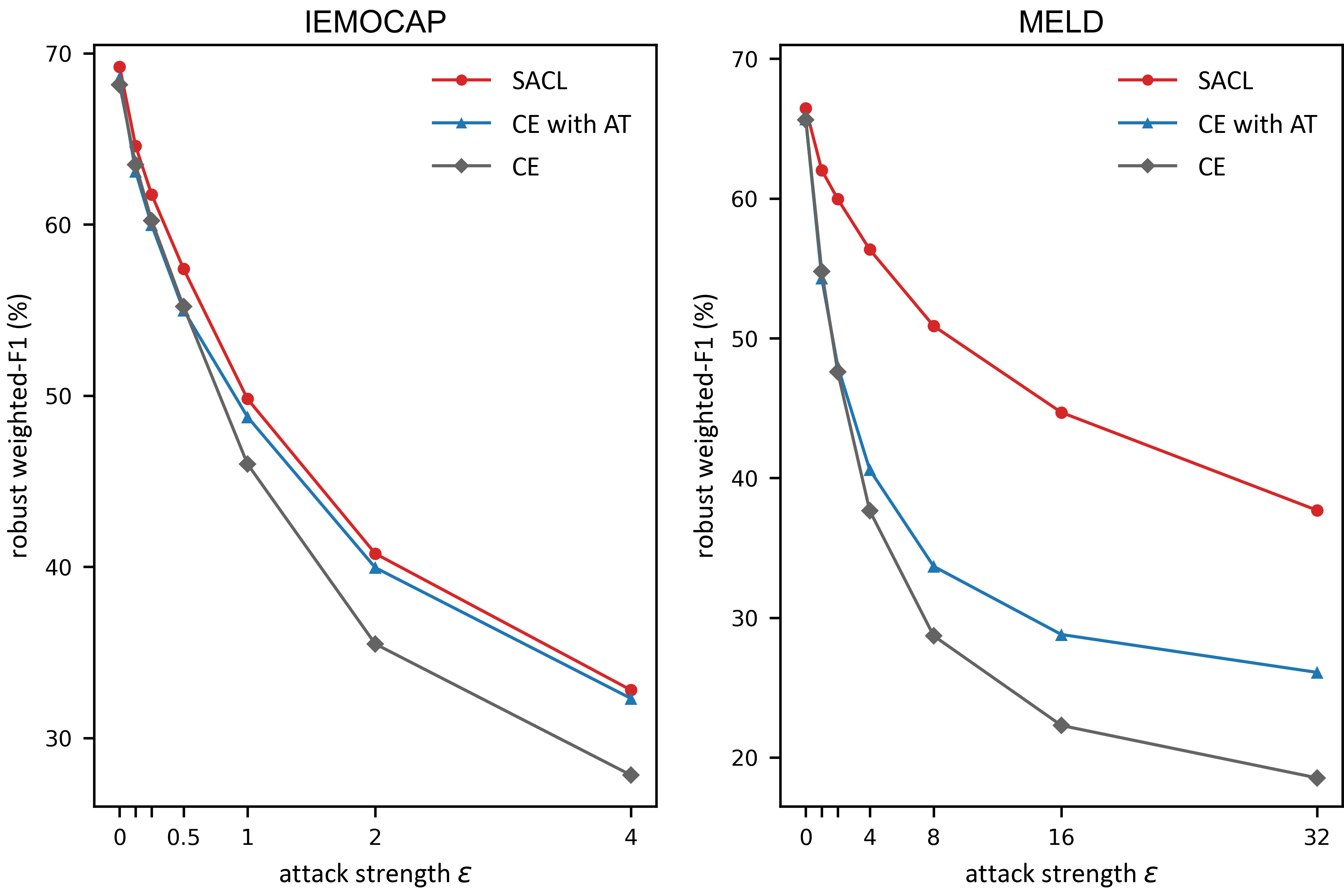}
    \caption{Context robustness performances against different optimization objectives. We report the robust weighted-F1 scores under different attack strengths.
    Detailed results are listed in Appendix~\ref{sec:app:robust}.
    }
  \label{fig:robust}
\end{figure}

\subsection{Context Robustness Evaluation}
We further validate context robustness against different optimization objectives. We adjust different attack strengths of CE-based contextual adversarial perturbations on the test set and report the robust weighted-F1 scores.
The context robustness results of SACL, CE with AT, and CE objectives  on IEMOCAP and MELD are shown in Figure~\ref{fig:robust}.
CE with AT means using a cross-entropy objective with traditional adversarial training, i.e., FGM.

Our SACL consistently gains better robust weighted-F1 scores over other optimization objectives on both datasets. 
Under different attack strengths ($\epsilon>0$), SACL-LSTM achieves up to \textbf{2.2\%} (average  \textbf{1.3\%}) and \textbf{17.2\%} (average \textbf{13.4\%}) absolute improvements on IEMOCAP and MELD, respectively.
CE with AT obtains sub-optimal performance 
since generating context-free adversarial samples interferes with the correlation between utterances, which is detrimental to context understanding.
Our SACL using CAT can generate context-level worst-case samples for better training and enhance the model's context robustness.

Moreover, we observe that SACL achieves a significant improvement on MELD with limited context.
The average number of dialogue turns in MELD is relatively small, making it more likely for any two utterances to be strongly correlated. 
By introducing CAT, SACL learns more diverse features from the limited context, obtaining better context robustness results on MELD than others.

\begin{figure}[t]
  \centering
    \includegraphics[width=0.96\linewidth]{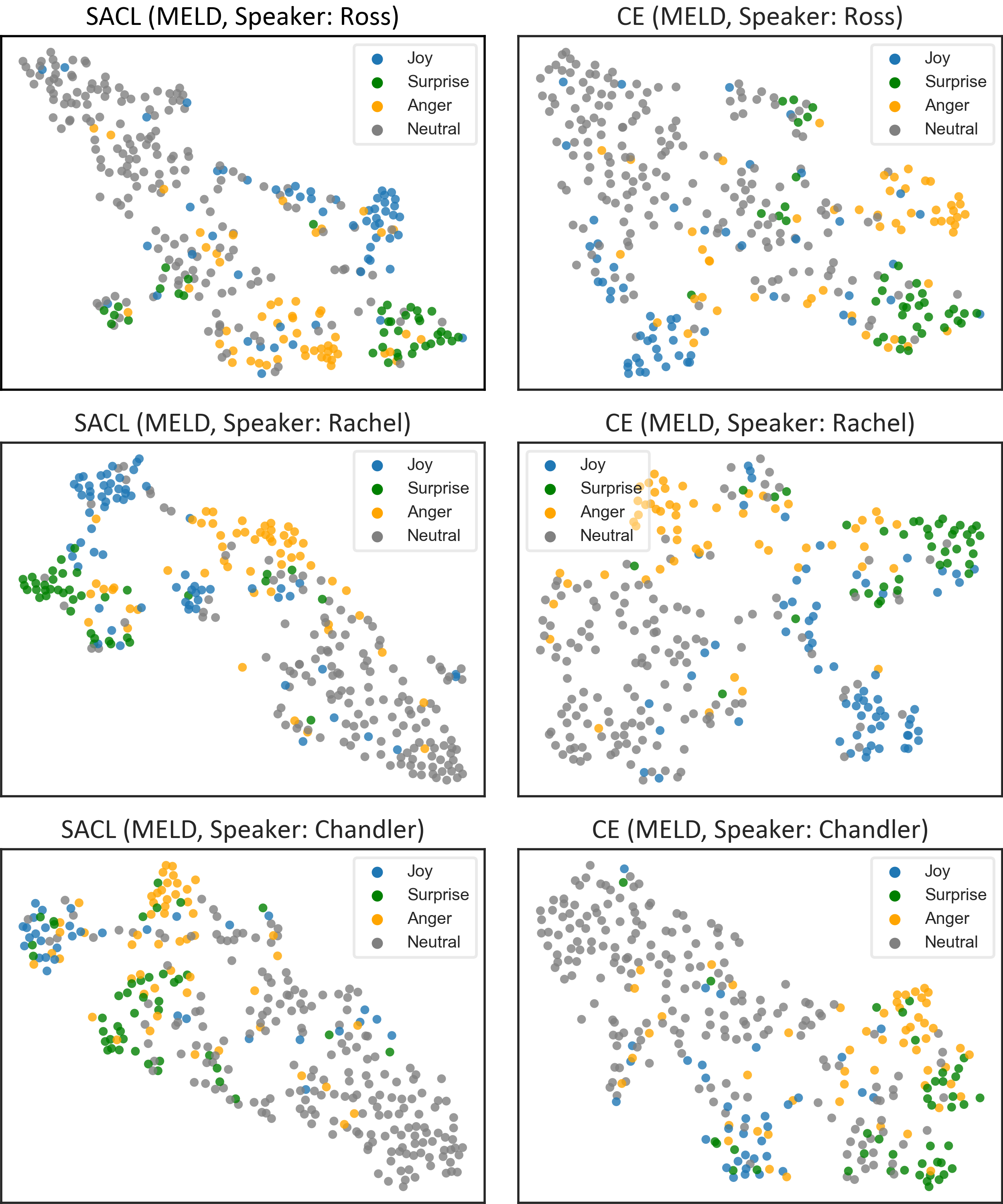}
    \caption{
    t-SNE visualization of representations learned with different optimization objectives on MELD. The data points reflect an overall distribution representation.
    Points corresponding to categories with a sample proportion of less than 10\% are excluded for a clear picture. 
    }
  \label{fig:vis}
\end{figure}

\begin{figure}
  \centering
    \includegraphics[width=\linewidth]{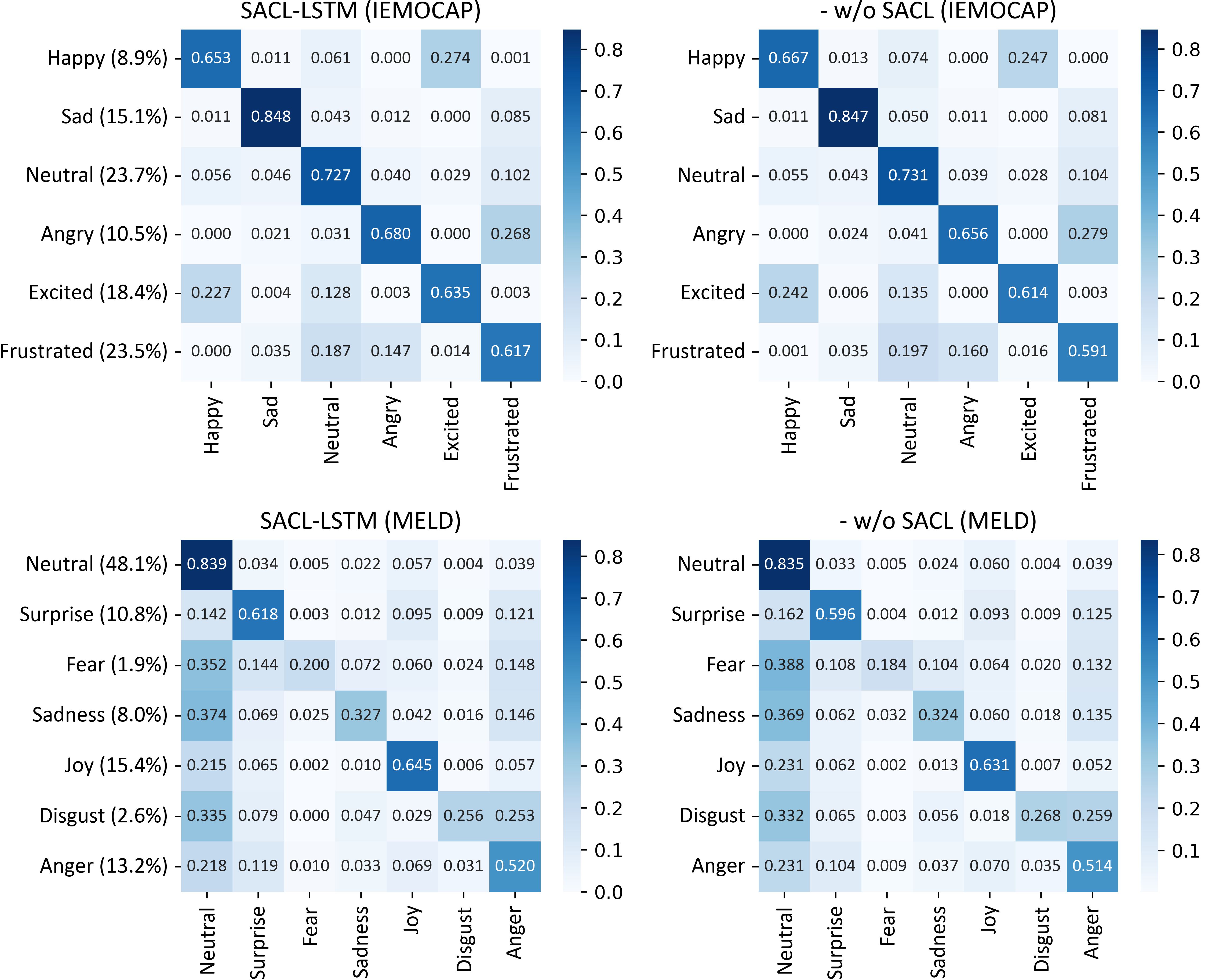}
    \caption{The normalized confusion matrices for SACL-LSTM and its variant.
    The rows represent the actual classes, whereas the columns represent predictions made by the model. 
    Each cell $(i,j)$ represents the percentage of class $i$ predicted to be class $j$.
    }
  \label{fig:confusion}
\end{figure}

\subsection{Representation Visualization}
We qualitatively visualize the learned representations on the test set of MELD with t-SNE \cite{van2008visualizing}. 
Figure~\ref{fig:vis} shows the visualization of the three speakers. 
Compared with using CE objective, the distribution of each emotion class learned by our SACL is more tight and united.
It indicates that SACL can learn cluster-level structured representations and have a better ability to generalization. 
Besides, under SACL, the representations of surprise are away from neutral, and close to both joy and anger, which is consistent with the nature of surprise\footnote{Surprise is a non-neutral complex emotion that can be expressed with positive or negative valence \cite{DBLP:conf/acl/PoriaHMNCM19}.}.
It reveals that SACL can partly learn inter-class intrinsic structure in addition to intra-class feature consistency.

\subsection{Error Analysis}
Figure~\ref{fig:confusion} shows an error analysis of SACL-LSTM and its ablated variant on the test set of IEMOCAP and MELD.
The normalized confusion matrices are used to evaluate the quality of each model's predicted outputs.
From the diagonal elements of the matrices, SACL-LSTM reports better true positives against others on most fine-grained emotion categories.
It suggests that SACL-LSTM is unbiased towards the under-represented emotion labels and learns better fine-grained features.
Compared with the ablated variant {{w/o SACL}}, SACL-LSTM obtains better performances at similar categories, e.g., excited to happy, angry to frustrated, and frustrated to angry on IEMOCAP. It indicates that the SACL framework can effectively mitigate the misclassification problem of similar emotions. 
The poor effect of happy to excited may be due to the small proportion of happy samples used for training.
For MELD, some categories (i.e., fear, sadness, and disgust) that account for a small proportion are easily misclassified as neutral accounting for nearly half, which is caused by the class imbalance issue.

\section{Conclusion}
We propose a supervised adversarial contrastive learning framework to learn class-spread structured representations for classification. 
It applies a contrast-aware adversarial training strategy and a joint class-spread contrastive learning objective.
Besides, we design a contextual adversarial training strategy to learn more diverse features from context-dependent inputs and enhance the model's context robustness.
Under the SACL framework with CAT, we develop a sequence-based method SACL-LSTM to learn label-consistent and context-robust features on context-dependent data for better emotion recognition. 
Experiments verified the effectiveness of SACL-LSTM for ERC and SACL for learning generalized and robust representations.

\section*{Limitations}
In this paper, we present a supervised adversarial contrastive learning (SACL) framework with contextual adversarial training to learn class-spread structured representations for context-dependent emotion classification.
However, the framework is somewhat limited by the class imbalance issue, as illustrated in Section~\ref{sec:sec:exp}.
To more comprehensively evaluate the generalization of SACL, it is necessary to test its transferability in low-resource and out-of-distribution scenarios, and evaluate its performance across a wider range of tasks.
Additionally, it would be beneficial to explore the theoretical underpinnings and potential applications of the framework in greater depth.
The aforementioned limitations will be left for future research.

\section*{Acknowledgements}
This work was supported by the National Key Research and Development Program of China (No. 2022YFC3302102) and the National Natural Science Foundation of China (No. 62102412).
The authors thank the anonymous reviewers and the meta-reviewer for their helpful comments on the paper.

\bibliography{anthology,dialogue}

\begin{thebibliography}{70}
\expandafter\ifx\csname natexlab\endcsname\relax\def\natexlab#1{#1}\fi

\bibitem[{Bachman et~al.(2019)Bachman, Hjelm, and
  Buchwalter}]{DBLP:conf/nips/BachmanHB19}
Philip Bachman, R.~Devon Hjelm, and William Buchwalter. 2019.
\newblock \href
  {https://proceedings.neurips.cc/paper/2019/hash/ddf354219aac374f1d40b7e760ee5bb7-Abstract.html}
  {Learning representations by maximizing mutual information across views}.
\newblock In \emph{Advances in Neural Information Processing Systems 32: Annual
  Conference on Neural Information Processing Systems 2019, NeurIPS 2019,
  December 8-14, 2019, Vancouver, BC, Canada}, pages 15509--15519.

\bibitem[{Bao et~al.(2022)Bao, Ma, Wei, Zhou, and
  Hu}]{DBLP:conf/ijcai/BaoMWZH22}
Yinan Bao, Qianwen Ma, Lingwei Wei, Wei Zhou, and Songlin Hu. 2022.
\newblock \href {https://doi.org/10.24963/ijcai.2022/562} {Speaker-guided
  encoder-decoder framework for emotion recognition in conversation}.
\newblock In \emph{Proceedings of the Thirty-First International Joint
  Conference on Artificial Intelligence, {IJCAI} 2022, Vienna, Austria, 23-29
  July 2022}, pages 4051--4057. ijcai.org.

\bibitem[{Bengio et~al.(2009)Bengio, Louradour, Collobert, and
  Weston}]{DBLP:conf/icml/BengioLCW09}
Yoshua Bengio, J{\'{e}}r{\^{o}}me Louradour, Ronan Collobert, and Jason Weston.
  2009.
\newblock \href {https://doi.org/10.1145/1553374.1553380} {Curriculum
  learning}.
\newblock In \emph{Proceedings of the 26th Annual International Conference on
  Machine Learning, {ICML} 2009, Montreal, Quebec, Canada, June 14-18, 2009},
  volume 382 of \emph{{ACM} International Conference Proceeding Series}, pages
  41--48. {ACM}.

\bibitem[{Busso et~al.(2008)Busso, Bulut, Lee, Kazemzadeh, Mower, Kim, Chang,
  Lee, and Narayanan}]{DBLP:journals/lre/BussoBLKMKCLN08}
Carlos Busso, Murtaza Bulut, Chi{-}Chun Lee, Abe Kazemzadeh, Emily Mower,
  Samuel Kim, Jeannette~N. Chang, Sungbok Lee, and Shrikanth~S. Narayanan.
  2008.
\newblock \href {https://doi.org/10.1007/s10579-008-9076-6} {{IEMOCAP:}
  {I}nteractive emotional dyadic motion capture database}.
\newblock \emph{Lang. Resour. Evaluation}, 42(4):335--359.

\bibitem[{Carlini and Wagner(2017)}]{DBLP:conf/sp/Carlini017}
Nicholas Carlini and David~A. Wagner. 2017.
\newblock \href {https://doi.org/10.1109/SP.2017.49} {Towards evaluating the
  robustness of neural networks}.
\newblock In \emph{2017 {IEEE} Symposium on Security and Privacy, {SP} 2017,
  San Jose, CA, USA, May 22-26, 2017}, pages 39--57. {IEEE} Computer Society.

\bibitem[{Chen et~al.(2020)Chen, Kornblith, Norouzi, and
  Hinton}]{DBLP:conf/icml/ChenK0H20}
Ting Chen, Simon Kornblith, Mohammad Norouzi, and Geoffrey~E. Hinton. 2020.
\newblock \href {http://proceedings.mlr.press/v119/chen20j.html} {A simple
  framework for contrastive learning of visual representations}.
\newblock In \emph{Proceedings of the 37th International Conference on Machine
  Learning, {ICML} 2020, 13-18 July 2020, Virtual Event}, volume 119 of
  \emph{Proceedings of Machine Learning Research}, pages 1597--1607. {PMLR}.

\bibitem[{Chopra et~al.(2005)Chopra, Hadsell, and
  LeCun}]{DBLP:conf/cvpr/ChopraHL05}
Sumit Chopra, Raia Hadsell, and Yann LeCun. 2005.
\newblock \href {https://doi.org/10.1109/CVPR.2005.202} {Learning a similarity
  metric discriminatively, with application to face verification}.
\newblock In \emph{2005 {IEEE} Computer Society Conference on Computer Vision
  and Pattern Recognition {(CVPR} 2005), 20-26 June 2005, San Diego, CA,
  {USA}}, pages 539--546. {IEEE} Computer Society.

\bibitem[{Devlin et~al.(2019)Devlin, Chang, Lee, and
  Toutanova}]{DBLP:conf/naacl/DevlinCLT19}
Jacob Devlin, Ming{-}Wei Chang, Kenton Lee, and Kristina Toutanova. 2019.
\newblock \href {https://doi.org/10.18653/v1/n19-1423} {{BERT:} pre-training of
  deep bidirectional transformers for language understanding}.
\newblock In \emph{Proceedings of the 2019 Conference of the North American
  Chapter of the Association for Computational Linguistics: Human Language
  Technologies, {NAACL-HLT} 2019, Minneapolis, MN, USA, June 2-7, 2019, Volume
  1 (Long and Short Papers)}, pages 4171--4186. Association for Computational
  Linguistics.

\bibitem[{Fan et~al.(2021)Fan, Liu, Chen, Zhang, and
  Gan}]{DBLP:conf/nips/FanLCZG21}
Lijie Fan, Sijia Liu, Pin{-}Yu Chen, Gaoyuan Zhang, and Chuang Gan. 2021.
\newblock \href
  {https://proceedings.neurips.cc/paper/2021/hash/b36ed8a07e3cd80ee37138524690eca1-Abstract.html}
  {When does contrastive learning preserve adversarial robustness from
  pretraining to finetuning?}
\newblock In \emph{Advances in Neural Information Processing Systems 34: Annual
  Conference on Neural Information Processing Systems 2021, NeurIPS 2021,
  December 6-14, 2021, virtual}, pages 21480--21492.

\bibitem[{Ghosal et~al.(2020)Ghosal, Majumder, Gelbukh, Mihalcea, and
  Poria}]{DBLP:conf/emnlp/GhosalMGMP20}
Deepanway Ghosal, Navonil Majumder, Alexander~F. Gelbukh, Rada Mihalcea, and
  Soujanya Poria. 2020.
\newblock \href {https://doi.org/10.18653/v1/2020.findings-emnlp.224}
  {{COSMIC:} {C}ommonsense knowledge for emotion identification in
  conversations}.
\newblock In \emph{Findings of the Association for Computational Linguistics:
  {EMNLP} 2020, Online Event, 16-20 November 2020}, volume {EMNLP} 2020 of
  \emph{Findings of {ACL}}, pages 2470--2481. Association for Computational
  Linguistics.

\bibitem[{Ghosal et~al.(2019)Ghosal, Majumder, Poria, Chhaya, and
  Gelbukh}]{DBLP:conf/emnlp/GhosalMPCG19}
Deepanway Ghosal, Navonil Majumder, Soujanya Poria, Niyati Chhaya, and
  Alexander~F. Gelbukh. 2019.
\newblock \href {https://doi.org/10.18653/v1/D19-1015} {{DialogueGCN:} {A}
  graph convolutional neural network for emotion recognition in conversation}.
\newblock In \emph{Proceedings of the 2019 Conference on Empirical Methods in
  Natural Language Processing and the 9th International Joint Conference on
  Natural Language Processing, {EMNLP-IJCNLP} 2019, Hong Kong, China, November
  3-7, 2019}, pages 154--164. Association for Computational Linguistics.

\bibitem[{Goodfellow et~al.(2015)Goodfellow, Shlens, and
  Szegedy}]{DBLP:journals/corr/GoodfellowSS14}
Ian~J. Goodfellow, Jonathon Shlens, and Christian Szegedy. 2015.
\newblock \href {http://arxiv.org/abs/1412.6572} {Explaining and harnessing
  adversarial examples}.
\newblock In \emph{3rd International Conference on Learning Representations,
  {ICLR} 2015, San Diego, CA, USA, May 7-9, 2015, Conference Track
  Proceedings}.

\bibitem[{Graf et~al.(2021)Graf, Hofer, Niethammer, and
  Kwitt}]{graf2021dissecting}
Florian Graf, Christoph~D. Hofer, Marc Niethammer, and Roland Kwitt. 2021.
\newblock \href {http://proceedings.mlr.press/v139/graf21a.html} {Dissecting
  supervised constrastive learning}.
\newblock In \emph{Proceedings of the 38th International Conference on Machine
  Learning, {ICML} 2021, 18-24 July 2021, Virtual Event}, volume 139 of
  \emph{Proceedings of Machine Learning Research}, pages 3821--3830. {PMLR}.

\bibitem[{Gunel et~al.(2021)Gunel, Du, Conneau, and
  Stoyanov}]{gunel2020supervised}
Beliz Gunel, Jingfei Du, Alexis Conneau, and Veselin Stoyanov. 2021.
\newblock \href {https://openreview.net/forum?id=cu7IUiOhujH} {Supervised
  contrastive learning for pre-trained language model fine-tuning}.
\newblock In \emph{9th International Conference on Learning Representations,
  {ICLR} 2021, Virtual Event, Austria, May 3-7, 2021}. OpenReview.net.

\bibitem[{Hazarika et~al.(2018{\natexlab{a}})Hazarika, Poria, Mihalcea,
  Cambria, and Zimmermann}]{DBLP:conf/emnlp/HazarikaPMCZ18}
Devamanyu Hazarika, Soujanya Poria, Rada Mihalcea, Erik Cambria, and Roger
  Zimmermann. 2018{\natexlab{a}}.
\newblock \href {https://doi.org/10.18653/v1/d18-1280} {{ICON:} {I}nteractive
  conversational memory network for multimodal emotion detection}.
\newblock In \emph{Proceedings of the 2018 Conference on Empirical Methods in
  Natural Language Processing, Brussels, Belgium, October 31 - November 4,
  2018}, pages 2594--2604. Association for Computational Linguistics.

\bibitem[{Hazarika et~al.(2018{\natexlab{b}})Hazarika, Poria, Zadeh, Cambria,
  Morency, and Zimmermann}]{DBLP:conf/naacl/HazarikaPZCMZ18}
Devamanyu Hazarika, Soujanya Poria, Amir Zadeh, Erik Cambria, Louis{-}Philippe
  Morency, and Roger Zimmermann. 2018{\natexlab{b}}.
\newblock \href {https://doi.org/10.18653/v1/n18-1193} {Conversational memory
  network for emotion recognition in dyadic dialogue videos}.
\newblock In \emph{Proceedings of the 2018 Conference of the North American
  Chapter of the Association for Computational Linguistics: Human Language
  Technologies, {NAACL-HLT} 2018, New Orleans, Louisiana, USA, June 1-6, 2018,
  Volume 1 (Long Papers)}, pages 2122--2132. Association for Computational
  Linguistics.

\bibitem[{H{\'{e}}naff(2020)}]{DBLP:conf/icml/Henaff20}
Olivier~J. H{\'{e}}naff. 2020.
\newblock \href {http://proceedings.mlr.press/v119/henaff20a.html}
  {Data-efficient image recognition with contrastive predictive coding}.
\newblock In \emph{Proceedings of the 37th International Conference on Machine
  Learning, {ICML} 2020, 13-18 July 2020, Virtual Event}, volume 119 of
  \emph{Proceedings of Machine Learning Research}, pages 4182--4192. {PMLR}.

\bibitem[{Hochreiter and Schmidhuber(1997)}]{DBLP:journals/neco/HochreiterS97}
Sepp Hochreiter and J{\"{u}}rgen Schmidhuber. 1997.
\newblock \href {https://doi.org/10.1162/neco.1997.9.8.1735} {Long short-term
  memory}.
\newblock \emph{Neural Comput.}, 9(8):1735--1780.

\bibitem[{Hu et~al.(2022{\natexlab{a}})Hu, Hou, Du, Zhou, Jiang, Mo, and
  Shi}]{DBLP:conf/emnlp/0001HDZJMS22}
Dou Hu, Xiaolong Hou, Xiyang Du, Mengyuan Zhou, Lianxin Jiang, Yang Mo, and
  Xiaofeng Shi. 2022{\natexlab{a}}.
\newblock \href {https://aclanthology.org/2022.findings-emnlp.468} {{VarMAE:}
  {P}re-training of variational masked autoencoder for domain-adaptive language
  understanding}.
\newblock In \emph{Findings of the Association for Computational Linguistics:
  {EMNLP} 2022, Abu Dhabi, United Arab Emirates, December 7-11, 2022}, pages
  6276--6286. Association for Computational Linguistics.

\bibitem[{Hu et~al.(2022{\natexlab{b}})Hu, Hou, Wei, Jiang, and
  Mo}]{DBLP:conf/icassp/HuHWJM22}
Dou Hu, Xiaolong Hou, Lingwei Wei, Lian{-}Xin Jiang, and Yang Mo.
  2022{\natexlab{b}}.
\newblock \href {https://doi.org/10.1109/ICASSP43922.2022.9747397} {{MM-DFN:}
  {M}ultimodal dynamic fusion network for emotion recognition in
  conversations}.
\newblock In \emph{{IEEE} International Conference on Acoustics, Speech and
  Signal Processing, {ICASSP} 2022, Virtual and Singapore, 23-27 May 2022},
  pages 7037--7041. {IEEE}.

\bibitem[{Hu et~al.(2021{\natexlab{a}})Hu, Wei, and
  Huai}]{DBLP:conf/acl/HuWH20}
Dou Hu, Lingwei Wei, and Xiaoyong Huai. 2021{\natexlab{a}}.
\newblock \href {https://doi.org/10.18653/v1/2021.acl-long.547} {{DialogueCRN:}
  {C}ontextual reasoning networks for emotion recognition in conversations}.
\newblock In \emph{Proceedings of the 59th Annual Meeting of the Association
  for Computational Linguistics and the 11th International Joint Conference on
  Natural Language Processing, {ACL/IJCNLP} 2021, (Volume 1: Long Papers),
  Virtual Event, August 1-6, 2021}, pages 7042--7052. Association for
  Computational Linguistics.

\bibitem[{Hu et~al.(2022{\natexlab{c}})Hu, Zhou, Du, Yuan, Zhi, Jiang, Mo, and
  Shi}]{DBLP:conf/semeval/0001ZDYZJMS22}
Dou Hu, Mengyuan Zhou, Xiyang Du, Mengfei Yuan, Jin Zhi, Lian{-}Xin Jiang, Yang
  Mo, and Xiaofeng Shi. 2022{\natexlab{c}}.
\newblock \href {https://doi.org/10.18653/v1/2022.semeval-1.43} {{PALI-NLP} at
  {SemEval-2022} {T}ask 4: {D}iscriminative fine-tuning of transformers for
  patronizing and condescending language detection}.
\newblock In \emph{Proceedings of the 16th International Workshop on Semantic
  Evaluation, SemEval@NAACL 2022, Seattle, Washington, United States, July
  14-15, 2022}, pages 335--343. Association for Computational Linguistics.

\bibitem[{Hu et~al.(2021{\natexlab{b}})Hu, Liu, Zhao, and
  Jin}]{DBLP:conf/acl/HuLZJ20}
Jingwen Hu, Yuchen Liu, Jinming Zhao, and Qin Jin. 2021{\natexlab{b}}.
\newblock \href {https://doi.org/10.18653/v1/2021.acl-long.440} {{MMGCN:}
  {M}ultimodal fusion via deep graph convolution network for emotion
  recognition in conversation}.
\newblock In \emph{Proceedings of the 59th Annual Meeting of the Association
  for Computational Linguistics and the 11th International Joint Conference on
  Natural Language Processing, {ACL/IJCNLP} 2021, (Volume 1: Long Papers),
  Virtual Event, August 1-6, 2021}, pages 5666--5675. Association for
  Computational Linguistics.

\bibitem[{Ishiwatari et~al.(2020)Ishiwatari, Yasuda, Miyazaki, and
  Goto}]{DBLP:conf/emnlp/IshiwatariYMG20}
Taichi Ishiwatari, Yuki Yasuda, Taro Miyazaki, and Jun Goto. 2020.
\newblock \href {https://doi.org/10.18653/v1/2020.emnlp-main.597}
  {Relation-aware graph attention networks with relational position encodings
  for emotion recognition in conversations}.
\newblock In \emph{Proceedings of the 2020 Conference on Empirical Methods in
  Natural Language Processing, {EMNLP} 2020, Online, November 16-20, 2020},
  pages 7360--7370. Association for Computational Linguistics.

\bibitem[{Jiang et~al.(2020{\natexlab{a}})Jiang, He, Chen, Liu, Gao, and
  Zhao}]{DBLP:conf/acl/JiangHCLGZ20}
Haoming Jiang, Pengcheng He, Weizhu Chen, Xiaodong Liu, Jianfeng Gao, and Tuo
  Zhao. 2020{\natexlab{a}}.
\newblock \href {https://doi.org/10.18653/v1/2020.acl-main.197} {{SMART:}
  {R}obust and efficient fine-tuning for pre-trained natural language models
  through principled regularized optimization}.
\newblock In \emph{Proceedings of the 58th Annual Meeting of the Association
  for Computational Linguistics, {ACL} 2020, Online, July 5-10, 2020}, pages
  2177--2190. Association for Computational Linguistics.

\bibitem[{Jiang et~al.(2020{\natexlab{b}})Jiang, Chen, Chen, and
  Wang}]{DBLP:conf/nips/JiangCCW20}
Ziyu Jiang, Tianlong Chen, Ting Chen, and Zhangyang Wang. 2020{\natexlab{b}}.
\newblock \href
  {https://proceedings.neurips.cc/paper/2020/hash/ba7e36c43aff315c00ec2b8625e3b719-Abstract.html}
  {Robust pre-training by adversarial contrastive learning}.
\newblock In \emph{Advances in Neural Information Processing Systems 33: Annual
  Conference on Neural Information Processing Systems 2020, NeurIPS 2020,
  December 6-12, 2020, virtual}.

\bibitem[{Jiao et~al.(2020{\natexlab{a}})Jiao, Lyu, and
  King}]{DBLP:conf/emnlp/JiaoLK20}
Wenxiang Jiao, Michael~R. Lyu, and Irwin King. 2020{\natexlab{a}}.
\newblock \href {https://doi.org/10.18653/v1/2020.findings-emnlp.435}
  {Exploiting unsupervised data for emotion recognition in conversations}.
\newblock In \emph{Findings of the Association for Computational Linguistics:
  {EMNLP} 2020, Online Event, 16-20 November 2020}, volume {EMNLP} 2020 of
  \emph{Findings of {ACL}}, pages 4839--4846. Association for Computational
  Linguistics.

\bibitem[{Jiao et~al.(2020{\natexlab{b}})Jiao, Lyu, and
  King}]{DBLP:conf/aaai/JiaoLK20}
Wenxiang Jiao, Michael~R. Lyu, and Irwin King. 2020{\natexlab{b}}.
\newblock \href {https://ojs.aaai.org/index.php/AAAI/article/view/6309}
  {Real-time emotion recognition via attention gated hierarchical memory
  network}.
\newblock In \emph{The Thirty-Fourth {AAAI} Conference on Artificial
  Intelligence, {AAAI} 2020, The Thirty-Second Innovative Applications of
  Artificial Intelligence Conference, {IAAI} 2020, The Tenth {AAAI} Symposium
  on Educational Advances in Artificial Intelligence, {EAAI} 2020, New York,
  NY, USA, February 7-12, 2020}, pages 8002--8009. {AAAI} Press.

\bibitem[{Khosla et~al.(2020)Khosla, Teterwak, Wang, Sarna, Tian, Isola,
  Maschinot, Liu, and Krishnan}]{khosla2020supervised}
Prannay Khosla, Piotr Teterwak, Chen Wang, Aaron Sarna, Yonglong Tian, Phillip
  Isola, Aaron Maschinot, Ce~Liu, and Dilip Krishnan. 2020.
\newblock \href
  {https://proceedings.neurips.cc/paper/2020/hash/d89a66c7c80a29b1bdbab0f2a1a94af8-Abstract.html}
  {Supervised contrastive learning}.
\newblock In \emph{Advances in Neural Information Processing Systems 33: Annual
  Conference on Neural Information Processing Systems 2020, NeurIPS 2020,
  December 6-12, 2020, virtual}.

\bibitem[{Kim et~al.(2020)Kim, Tack, and Hwang}]{DBLP:conf/nips/KimTH20}
Minseon Kim, Jihoon Tack, and Sung~Ju Hwang. 2020.
\newblock \href
  {https://proceedings.neurips.cc/paper/2020/hash/1f1baa5b8edac74eb4eaa329f14a0361-Abstract.html}
  {Adversarial self-supervised contrastive learning}.
\newblock In \emph{Advances in Neural Information Processing Systems 33: Annual
  Conference on Neural Information Processing Systems 2020, NeurIPS 2020,
  December 6-12, 2020, virtual}.

\bibitem[{Kim(2015)}]{kim2015t}
Tae~Kyun Kim. 2015.
\newblock T test as a parametric statistic.
\newblock \emph{Korean journal of anesthesiology}, 68(6):540--546.

\bibitem[{Kingma and Ba(2015)}]{DBLP:journals/corr/KingmaB14}
Diederik~P. Kingma and Jimmy Ba. 2015.
\newblock \href {http://arxiv.org/abs/1412.6980} {Adam: A method for stochastic
  optimization}.
\newblock In \emph{3rd International Conference on Learning Representations,
  {ICLR} 2015, San Diego, CA, USA, May 7-9, 2015, Conference Track
  Proceedings}.

\bibitem[{Lee and Choi(2021)}]{DBLP:conf/emnlp/LeeC21}
Bongseok Lee and Yong~Suk Choi. 2021.
\newblock \href {https://doi.org/10.18653/v1/2021.emnlp-main.36} {Graph based
  network with contextualized representations of turns in dialogue}.
\newblock In \emph{Proceedings of the 2021 Conference on Empirical Methods in
  Natural Language Processing, {EMNLP} 2021, Virtual Event / Punta Cana,
  Dominican Republic, 7-11 November, 2021}, pages 443--455. Association for
  Computational Linguistics.

\bibitem[{Lee and Lee(2022)}]{DBLP:conf/naacl/LeeL22}
Joosung Lee and Wooin Lee. 2022.
\newblock \href {https://doi.org/10.18653/v1/2022.naacl-main.416} {{CoMPM:}
  {C}ontext modeling with speaker's pre-trained memory tracking for emotion
  recognition in conversation}.
\newblock In \emph{Proceedings of the 2022 Conference of the North American
  Chapter of the Association for Computational Linguistics: Human Language
  Technologies, {NAACL} 2022, Seattle, WA, United States, July 10-15, 2022},
  pages 5669--5679. Association for Computational Linguistics.

\bibitem[{Li et~al.(2021{\natexlab{a}})Li, Lin, Fu, and
  Wang}]{DBLP:conf/emnlp/Li00W21}
Jiangnan Li, Zheng Lin, Peng Fu, and Weiping Wang. 2021{\natexlab{a}}.
\newblock \href {https://doi.org/10.18653/v1/2021.findings-emnlp.104} {Past,
  present, and future: Conversational emotion recognition through structural
  modeling of psychological knowledge}.
\newblock In \emph{Findings of the Association for Computational Linguistics:
  {EMNLP} 2021, Virtual Event / Punta Cana, Dominican Republic, 16-20 November,
  2021}, pages 1204--1214. Association for Computational Linguistics.

\bibitem[{Li et~al.(2021{\natexlab{b}})Li, Zhou, Xiong, and
  Hoi}]{DBLP:conf/iclr/0001ZXH21}
Junnan Li, Pan Zhou, Caiming Xiong, and Steven C.~H. Hoi. 2021{\natexlab{b}}.
\newblock \href {https://openreview.net/forum?id=KmykpuSrjcq} {Prototypical
  contrastive learning of unsupervised representations}.
\newblock In \emph{9th International Conference on Learning Representations,
  {ICLR} 2021, Virtual Event, Austria, May 3-7, 2021}. OpenReview.net.

\bibitem[{Li et~al.(2022{\natexlab{a}})Li, Yan, and
  Qiu}]{DBLP:conf/aaai/LiYQ22}
Shimin Li, Hang Yan, and Xipeng Qiu. 2022{\natexlab{a}}.
\newblock \href {https://ojs.aaai.org/index.php/AAAI/article/view/21348}
  {Contrast and generation make {BART} a good dialogue emotion recognizer}.
\newblock In \emph{Thirty-Sixth {AAAI} Conference on Artificial Intelligence,
  {AAAI} 2022, Thirty-Fourth Conference on Innovative Applications of
  Artificial Intelligence, {IAAI} 2022, The Twelveth Symposium on Educational
  Advances in Artificial Intelligence, {EAAI} 2022 Virtual Event, February 22 -
  March 1, 2022}, pages 11002--11010. {AAAI} Press.

\bibitem[{Li et~al.(2022{\natexlab{b}})Li, Zhou, Zhang, Liu, Yang, Lian, and
  Hu}]{DBLP:conf/coling/LiZZLYLH22}
Ziming Li, Yan Zhou, Weibo Zhang, Yaxin Liu, Chuanpeng Yang, Zheng Lian, and
  Songlin Hu. 2022{\natexlab{b}}.
\newblock \href {https://aclanthology.org/2022.coling-1.623} {{AMOA:} global
  acoustic feature enhanced modal-order-aware network for multimodal sentiment
  analysis}.
\newblock In \emph{Proceedings of the 29th International Conference on
  Computational Linguistics, {COLING} 2022, Gyeongju, Republic of Korea,
  October 12-17, 2022}, pages 7136--7146. International Committee on
  Computational Linguistics.

\bibitem[{Lin et~al.(2022)Lin, Ma, Chen, Yang, Cheng, and
  Chen}]{DBLP:conf/naacl/LinMCYCC22}
Hongzhan Lin, Jing Ma, Liangliang Chen, Zhiwei Yang, Mingfei Cheng, and Guang
  Chen. 2022.
\newblock \href {https://doi.org/10.18653/v1/2022.findings-naacl.194} {Detect
  rumors in microblog posts for low-resource domains via adversarial
  contrastive learning}.
\newblock In \emph{Findings of the Association for Computational Linguistics:
  {NAACL} 2022, Seattle, WA, United States, July 10-15, 2022}, pages
  2543--2556. Association for Computational Linguistics.

\bibitem[{Lin et~al.(2017)Lin, Goyal, Girshick, He, and
  Doll{\'{a}}r}]{DBLP:conf/iccv/LinGGHD17}
Tsung{-}Yi Lin, Priya Goyal, Ross~B. Girshick, Kaiming He, and Piotr
  Doll{\'{a}}r. 2017.
\newblock \href {https://doi.org/10.1109/ICCV.2017.324} {Focal loss for dense
  object detection}.
\newblock In \emph{{IEEE} International Conference on Computer Vision, {ICCV}
  2017, Venice, Italy, October 22-29, 2017}, pages 2999--3007. {IEEE} Computer
  Society.

\bibitem[{Liu et~al.(2019)Liu, Ott, Goyal, Du, Joshi, Chen, Levy, Lewis,
  Zettlemoyer, and Stoyanov}]{DBLP:journals/corr/abs-1907-11692}
Yinhan Liu, Myle Ott, Naman Goyal, Jingfei Du, Mandar Joshi, Danqi Chen, Omer
  Levy, Mike Lewis, Luke Zettlemoyer, and Veselin Stoyanov. 2019.
\newblock \href {http://arxiv.org/abs/1907.11692} {{RoBERTa:} {A} robustly
  optimized {BERT} pretraining approach}.
\newblock \emph{CoRR}, abs/1907.11692.

\bibitem[{Ma et~al.(2020)Ma, Nguyen, Xing, and
  Cambria}]{DBLP:journals/inffus/MaNXC20}
Yukun Ma, Khanh~Linh Nguyen, Frank~Z. Xing, and Erik Cambria. 2020.
\newblock \href {https://doi.org/10.1016/j.inffus.2020.06.011} {A survey on
  empathetic dialogue systems}.
\newblock \emph{Inf. Fusion}, 64:50--70.

\bibitem[{Majumder et~al.(2019)Majumder, Poria, Hazarika, Mihalcea, Gelbukh,
  and Cambria}]{DBLP:conf/aaai/MajumderPHMGC19}
Navonil Majumder, Soujanya Poria, Devamanyu Hazarika, Rada Mihalcea,
  Alexander~F. Gelbukh, and Erik Cambria. 2019.
\newblock \href {https://doi.org/10.1609/aaai.v33i01.33016818} {{DialogueRNN}:
  {A}n attentive {RNN} for emotion detection in conversations}.
\newblock In \emph{The Thirty-Third {AAAI} Conference on Artificial
  Intelligence, {AAAI} 2019, The Thirty-First Innovative Applications of
  Artificial Intelligence Conference, {IAAI} 2019, The Ninth {AAAI} Symposium
  on Educational Advances in Artificial Intelligence, {EAAI} 2019, Honolulu,
  Hawaii, USA, January 27 - February 1, 2019}, pages 6818--6825. {AAAI} Press.

\bibitem[{Mart{\'{\i}}n et~al.(2022)Mart{\'{\i}}n, S{\'{a}}nchez{-}Esguevillas,
  Arribas, and Carro}]{lopez2022supervised}
Manuel~L{\'{o}}pez Mart{\'{\i}}n, Antonio S{\'{a}}nchez{-}Esguevillas,
  Juan~Ignacio Arribas, and Bel{\'{e}}n Carro. 2022.
\newblock \href {https://doi.org/10.1016/j.inffus.2021.09.014} {Supervised
  contrastive learning over prototype-label embeddings for network intrusion
  detection}.
\newblock \emph{Inf. Fusion}, 79:200--228.

\bibitem[{Miyato et~al.(2017)Miyato, Dai, and
  Goodfellow}]{DBLP:conf/iclr/MiyatoDG17}
Takeru Miyato, Andrew~M. Dai, and Ian~J. Goodfellow. 2017.
\newblock \href {https://openreview.net/forum?id=r1X3g2\_xl} {Adversarial
  training methods for semi-supervised text classification}.
\newblock In \emph{5th International Conference on Learning Representations,
  {ICLR} 2017, Toulon, France, April 24-26, 2017, Conference Track
  Proceedings}. OpenReview.net.

\bibitem[{Poria et~al.(2017)Poria, Cambria, Hazarika, Majumder, Zadeh, and
  Morency}]{DBLP:conf/acl/PoriaCHMZM17}
Soujanya Poria, Erik Cambria, Devamanyu Hazarika, Navonil Majumder, Amir Zadeh,
  and Louis{-}Philippe Morency. 2017.
\newblock \href {https://doi.org/10.18653/v1/P17-1081} {Context-dependent
  sentiment analysis in user-generated videos}.
\newblock In \emph{Proceedings of the 55th Annual Meeting of the Association
  for Computational Linguistics, {ACL} 2017, Vancouver, Canada, July 30 -
  August 4, Volume 1: Long Papers}, pages 873--883. Association for
  Computational Linguistics.

\bibitem[{Poria et~al.(2019{\natexlab{a}})Poria, Hazarika, Majumder, Naik,
  Cambria, and Mihalcea}]{DBLP:conf/acl/PoriaHMNCM19}
Soujanya Poria, Devamanyu Hazarika, Navonil Majumder, Gautam Naik, Erik
  Cambria, and Rada Mihalcea. 2019{\natexlab{a}}.
\newblock \href {https://doi.org/10.18653/v1/p19-1050} {{MELD:} a multimodal
  multi-party dataset for emotion recognition in conversations}.
\newblock In \emph{Proceedings of the 57th Conference of the Association for
  Computational Linguistics, {ACL} 2019, Florence, Italy, July 28- August 2,
  2019, Volume 1: Long Papers}, pages 527--536. Association for Computational
  Linguistics.

\bibitem[{Poria et~al.(2019{\natexlab{b}})Poria, Majumder, Mihalcea, and
  Hovy}]{DBLP:journals/access/PoriaMMH19}
Soujanya Poria, Navonil Majumder, Rada Mihalcea, and Eduard~H. Hovy.
  2019{\natexlab{b}}.
\newblock \href {https://doi.org/10.1109/ACCESS.2019.2929050} {Emotion
  recognition in conversation: Research challenges, datasets, and recent
  advances}.
\newblock \emph{{IEEE} Access}, 7:100943--100953.

\bibitem[{Qin et~al.(2019)Qin, Martens, Gowal, Krishnan, Dvijotham, Fawzi, De,
  Stanforth, and Kohli}]{DBLP:conf/nips/QinMGKDFDSK19}
Chongli Qin, James Martens, Sven Gowal, Dilip Krishnan, Krishnamurthy
  Dvijotham, Alhussein Fawzi, Soham De, Robert Stanforth, and Pushmeet Kohli.
  2019.
\newblock \href
  {https://proceedings.neurips.cc/paper/2019/hash/0defd533d51ed0a10c5c9dbf93ee78a5-Abstract.html}
  {Adversarial robustness through local linearization}.
\newblock In \emph{Advances in Neural Information Processing Systems 32: Annual
  Conference on Neural Information Processing Systems 2019, NeurIPS 2019,
  December 8-14, 2019, Vancouver, BC, Canada}, pages 13824--13833.

\bibitem[{Shafahi et~al.(2019)Shafahi, Najibi, Ghiasi, Xu, Dickerson, Studer,
  Davis, Taylor, and Goldstein}]{DBLP:conf/nips/ShafahiNG0DSDTG19}
Ali Shafahi, Mahyar Najibi, Amin Ghiasi, Zheng Xu, John~P. Dickerson, Christoph
  Studer, Larry~S. Davis, Gavin Taylor, and Tom Goldstein. 2019.
\newblock \href
  {https://proceedings.neurips.cc/paper/2019/hash/7503cfacd12053d309b6bed5c89de212-Abstract.html}
  {Adversarial training for free!}
\newblock In \emph{Advances in Neural Information Processing Systems 32: Annual
  Conference on Neural Information Processing Systems 2019, NeurIPS 2019,
  December 8-14, 2019, Vancouver, BC, Canada}, pages 3353--3364.

\bibitem[{Shen et~al.(2021{\natexlab{a}})Shen, Chen, Quan, and
  Xie}]{DBLP:conf/aaai/ShenCQX21}
Weizhou Shen, Junqing Chen, Xiaojun Quan, and Zhixian Xie. 2021{\natexlab{a}}.
\newblock \href {https://ojs.aaai.org/index.php/AAAI/article/view/17625}
  {{DialogXL:} all-in-one xlnet for multi-party conversation emotion
  recognition}.
\newblock In \emph{Thirty-Fifth {AAAI} Conference on Artificial Intelligence,
  {AAAI} 2021, Thirty-Third Conference on Innovative Applications of Artificial
  Intelligence, {IAAI} 2021, The Eleventh Symposium on Educational Advances in
  Artificial Intelligence, {EAAI} 2021, Virtual Event, February 2-9, 2021},
  pages 13789--13797. {AAAI} Press.

\bibitem[{Shen et~al.(2021{\natexlab{b}})Shen, Wu, Yang, and
  Quan}]{DBLP:conf/acl/ShenWYQ20}
Weizhou Shen, Siyue Wu, Yunyi Yang, and Xiaojun Quan. 2021{\natexlab{b}}.
\newblock \href {https://doi.org/10.18653/v1/2021.acl-long.123} {Directed
  acyclic graph network for conversational emotion recognition}.
\newblock In \emph{Proceedings of the 59th Annual Meeting of the Association
  for Computational Linguistics and the 11th International Joint Conference on
  Natural Language Processing, {ACL/IJCNLP} 2021, (Volume 1: Long Papers),
  Virtual Event, August 1-6, 2021}, pages 1551--1560. Association for
  Computational Linguistics.

\bibitem[{Sohn(2016)}]{DBLP:conf/nips/Sohn16}
Kihyuk Sohn. 2016.
\newblock \href
  {https://proceedings.neurips.cc/paper/2016/hash/6b180037abbebea991d8b1232f8a8ca9-Abstract.html}
  {Improved deep metric learning with multi-class n-pair loss objective}.
\newblock In \emph{Advances in Neural Information Processing Systems 29: Annual
  Conference on Neural Information Processing Systems 2016, December 5-10,
  2016, Barcelona, Spain}, pages 1849--1857.

\bibitem[{Song et~al.(2022)Song, Huang, Xue, and Hu}]{DBLP:conf/emnlp/SongXH22}
Xiaohui Song, Longtao Huang, Hui Xue, and Songlin Hu. 2022.
\newblock \href {https://aclanthology.org/2022.emnlp-main.347} {Supervised
  prototypical contrastive learning for emotion recognition in conversation}.
\newblock In \emph{Proceedings of the 2022 Conference on Empirical Methods in
  Natural Language Processing, {EMNLP} 2022, Abu Dhabi, United Arab Emirates,
  December 7-11, 2022}, pages 5197--5206. Association for Computational
  Linguistics.

\bibitem[{Szegedy et~al.(2014)Szegedy, Zaremba, Sutskever, Bruna, Erhan,
  Goodfellow, and Fergus}]{DBLP:journals/corr/SzegedyZSBEGF13}
Christian Szegedy, Wojciech Zaremba, Ilya Sutskever, Joan Bruna, Dumitru Erhan,
  Ian~J. Goodfellow, and Rob Fergus. 2014.
\newblock \href {http://arxiv.org/abs/1312.6199} {Intriguing properties of
  neural networks}.
\newblock In \emph{2nd International Conference on Learning Representations,
  {ICLR} 2014, Banff, AB, Canada, April 14-16, 2014, Conference Track
  Proceedings}.

\bibitem[{Tian et~al.(2020)Tian, Krishnan, and Isola}]{DBLP:conf/eccv/TianKI20}
Yonglong Tian, Dilip Krishnan, and Phillip Isola. 2020.
\newblock \href {https://doi.org/10.1007/978-3-030-58621-8\_45} {Contrastive
  multiview coding}.
\newblock In \emph{Computer Vision - {ECCV} 2020 - 16th European Conference,
  Glasgow, UK, August 23-28, 2020, Proceedings, Part {XI}}, volume 12356 of
  \emph{Lecture Notes in Computer Science}, pages 776--794. Springer.

\bibitem[{van~den Oord et~al.(2018)van~den Oord, Li, and
  Vinyals}]{DBLP:journals/corr/abs-1807-03748}
A{\"{a}}ron van~den Oord, Yazhe Li, and Oriol Vinyals. 2018.
\newblock \href {http://arxiv.org/abs/1807.03748} {Representation learning with
  contrastive predictive coding}.
\newblock \emph{CoRR}, abs/1807.03748.

\bibitem[{Van~der Maaten and Hinton(2008)}]{van2008visualizing}
Laurens Van~der Maaten and Geoffrey Hinton. 2008.
\newblock Visualizing data using t-sne.
\newblock \emph{Journal of machine learning research}, 9(11).

\bibitem[{Wang et~al.(2021)Wang, Han, Wei, Zhang, and
  Wang}]{DBLP:conf/cvpr/00230W0W21}
Peng Wang, Kai Han, Xiu{-}Shen Wei, Lei Zhang, and Lei Wang. 2021.
\newblock \href {https://doi.org/10.1109/CVPR46437.2021.00100} {Contrastive
  learning based hybrid networks for long-tailed image classification}.
\newblock In \emph{{IEEE} Conference on Computer Vision and Pattern
  Recognition, {CVPR} 2021, virtual, June 19-25, 2021}, pages 943--952.
  Computer Vision Foundation / {IEEE}.

\bibitem[{Wang et~al.(2020)Wang, Zhang, Ma, Wang, and
  Xiao}]{DBLP:conf/sigdial/WangZMWX20}
Yan Wang, Jiayu Zhang, Jun Ma, Shaojun Wang, and Jing Xiao. 2020.
\newblock \href {https://aclanthology.org/2020.sigdial-1.23/} {Contextualized
  emotion recognition in conversation as sequence tagging}.
\newblock In \emph{Proceedings of the 21th Annual Meeting of the Special
  Interest Group on Discourse and Dialogue, SIGdial 2020, 1st virtual meeting,
  July 1-3, 2020}, pages 186--195. Association for Computational Linguistics.

\bibitem[{Wei et~al.(2020)Wei, Hu, Zhou, Tang, Zhang, Wang, Han, and
  Hu}]{wei2020hierarchical}
Lingwei Wei, Dou Hu, Wei Zhou, Xuehai Tang, Xiaodan Zhang, Xin Wang, Jizhong
  Han, and Songlin Hu. 2020.
\newblock \href {https://doi.org/10.1007/978-3-030-67664-3\_38} {Hierarchical
  interaction networks with rethinking mechanism for document-level sentiment
  analysis}.
\newblock In \emph{Machine Learning and Knowledge Discovery in Databases -
  European Conference, {ECML} {PKDD} 2020, Ghent, Belgium, September 14-18,
  2020, Proceedings, Part {III}}, volume 12459 of \emph{Lecture Notes in
  Computer Science}, pages 633--649. Springer.

\bibitem[{Yang et~al.(2022)Yang, Shen, Mao, and Cai}]{DBLP:conf/aaai/YangSMC22}
Lin Yang, Yi~Shen, Yue Mao, and Longjun Cai. 2022.
\newblock \href {https://ojs.aaai.org/index.php/AAAI/article/view/21413}
  {Hybrid curriculum learning for emotion recognition in conversation}.
\newblock In \emph{Thirty-Sixth {AAAI} Conference on Artificial Intelligence,
  {AAAI} 2022, Thirty-Fourth Conference on Innovative Applications of
  Artificial Intelligence, {IAAI} 2022, The Twelveth Symposium on Educational
  Advances in Artificial Intelligence, {EAAI} 2022 Virtual Event, February 22 -
  March 1, 2022}, pages 11595--11603. {AAAI} Press.

\bibitem[{Zahiri and Choi(2018)}]{DBLP:conf/aaai/ZahiriC18}
Sayyed~M. Zahiri and Jinho~D. Choi. 2018.
\newblock \href {https://aaai.org/ocs/index.php/WS/AAAIW18/paper/view/16434}
  {Emotion detection on {TV} show transcripts with sequence-based convolutional
  neural networks}.
\newblock In \emph{The Workshops of the The Thirty-Second {AAAI} Conference on
  Artificial Intelligence, New Orleans, Louisiana, USA, February 2-7, 2018},
  volume {WS-18} of \emph{{AAAI} Technical Report}, pages 44--52. {AAAI} Press.

\bibitem[{Zhang et~al.(2019{\natexlab{a}})Zhang, Zhang, Lu, Zhu, and
  Dong}]{DBLP:conf/nips/ZhangZLZ019}
Dinghuai Zhang, Tianyuan Zhang, Yiping Lu, Zhanxing Zhu, and Bin Dong.
  2019{\natexlab{a}}.
\newblock \href
  {https://proceedings.neurips.cc/paper/2019/hash/812b4ba287f5ee0bc9d43bbf5bbe87fb-Abstract.html}
  {You only propagate once: Accelerating adversarial training via maximal
  principle}.
\newblock In \emph{Advances in Neural Information Processing Systems 32: Annual
  Conference on Neural Information Processing Systems 2019, NeurIPS 2019,
  December 8-14, 2019, Vancouver, BC, Canada}, pages 227--238.

\bibitem[{Zhang et~al.(2019{\natexlab{b}})Zhang, Wu, Sun, Li, Zhu, and
  Zhou}]{DBLP:conf/ijcai/ZhangWSLZZ19}
Dong Zhang, Liangqing Wu, Changlong Sun, Shoushan Li, Qiaoming Zhu, and Guodong
  Zhou. 2019{\natexlab{b}}.
\newblock \href {https://doi.org/10.24963/ijcai.2019/752} {Modeling both
  context- and speaker-sensitive dependence for emotion detection in
  multi-speaker conversations}.
\newblock In \emph{Proceedings of the Twenty-Eighth International Joint
  Conference on Artificial Intelligence, {IJCAI} 2019, Macao, China, August
  10-16, 2019}, pages 5415--5421. ijcai.org.

\bibitem[{Zhao et~al.(2022)Zhao, Zhao, and Lu}]{DBLP:conf/ijcai/ZhaoZL22}
Weixiang Zhao, Yanyan Zhao, and Xin Lu. 2022.
\newblock \href {https://doi.org/10.24963/ijcai.2022/628} {{CauAIN}: Causal
  aware interaction network for emotion recognition in conversations}.
\newblock In \emph{Proceedings of the Thirty-First International Joint
  Conference on Artificial Intelligence, {IJCAI} 2022, Vienna, Austria, 23-29
  July 2022}, pages 4524--4530. ijcai.org.

\bibitem[{Zhong et~al.(2019)Zhong, Wang, and Miao}]{DBLP:conf/emnlp/ZhongWM19}
Peixiang Zhong, Di~Wang, and Chunyan Miao. 2019.
\newblock \href {https://doi.org/10.18653/v1/D19-1016} {Knowledge-enriched
  transformer for emotion detection in textual conversations}.
\newblock In \emph{Proceedings of the 2019 Conference on Empirical Methods in
  Natural Language Processing and the 9th International Joint Conference on
  Natural Language Processing, {EMNLP-IJCNLP} 2019, Hong Kong, China, November
  3-7, 2019}, pages 165--176. Association for Computational Linguistics.

\bibitem[{Zhou et~al.(2019)Zhou, Huang, Guo, Han, and
  Hu}]{DBLP:conf/ijcai/ZhouHGHH19}
Yan Zhou, Longtao Huang, Tao Guo, Jizhong Han, and Songlin Hu. 2019.
\newblock \href {https://doi.org/10.24963/ijcai.2019/762} {A span-based joint
  model for opinion target extraction and target sentiment classification}.
\newblock In \emph{Proceedings of the Twenty-Eighth International Joint
  Conference on Artificial Intelligence, {IJCAI} 2019, Macao, China, August
  10-16, 2019}, pages 5485--5491. ijcai.org.

\bibitem[{Zhu et~al.(2020)Zhu, Cheng, Gan, Sun, Goldstein, and
  Liu}]{DBLP:conf/iclr/ZhuCGSGL20}
Chen Zhu, Yu~Cheng, Zhe Gan, Siqi Sun, Tom Goldstein, and Jingjing Liu. 2020.
\newblock \href {https://openreview.net/forum?id=BygzbyHFvB} {Freelb: Enhanced
  adversarial training for natural language understanding}.
\newblock In \emph{8th International Conference on Learning Representations,
  {ICLR} 2020, Addis Ababa, Ethiopia, April 26-30, 2020}. OpenReview.net.

\bibitem[{Zhu et~al.(2021)Zhu, Pergola, Gui, Zhou, and
  He}]{DBLP:conf/acl/ZhuP0ZH20}
Lixing Zhu, Gabriele Pergola, Lin Gui, Deyu Zhou, and Yulan He. 2021.
\newblock \href {https://doi.org/10.18653/v1/2021.acl-long.125} {Topic-driven
  and knowledge-aware transformer for dialogue emotion detection}.
\newblock In \emph{Proceedings of the 59th Annual Meeting of the Association
  for Computational Linguistics and the 11th International Joint Conference on
  Natural Language Processing, {ACL/IJCNLP} 2021, (Volume 1: Long Papers),
  Virtual Event, August 1-6, 2021}, pages 1571--1582. Association for
  Computational Linguistics.

\end{thebibliography}
\bibliographystyle{acl_natbib}

\clearpage
\appendix

\section*{Appendix Overview}
In this supplementary material, we provide: 
(i) the related work, 
(ii) a detailed description of experimental setups,
and (iii) detailed results.

\section{Related Work}
\subsection{Emotion Recognition in Conversations}
Unlike traditional sentiment analysis \citep{DBLP:conf/ijcai/ZhouHGHH19,wei2020hierarchical,DBLP:conf/semeval/0001ZDYZJMS22,DBLP:conf/coling/LiZZLYLH22}, context information plays a significant role in identifying the emotion in conversations \cite{DBLP:journals/access/PoriaMMH19}.
Existing works usually 
utilize deep learning techniques to 
identify the emotion by context modeling and emotion representation learning.
These works can be roughly divided into sequence-, graph- and Transformer-based methods.

\subsubsection{Sequence-based Methods}
Sequence-based methods 
\cite{DBLP:conf/acl/PoriaCHMZM17,DBLP:conf/naacl/HazarikaPZCMZ18,DBLP:conf/emnlp/HazarikaPMCZ18,DBLP:conf/aaai/MajumderPHMGC19,DBLP:conf/emnlp/GhosalMGMP20,DBLP:conf/emnlp/JiaoLK20,DBLP:conf/aaai/JiaoLK20,DBLP:conf/acl/HuWH20,DBLP:conf/ijcai/ZhaoZL22} generally utilize sequential information in a dialogue to capture different levels of contextual features, i.e., situation, speakers and emotions. 
For example, \citet{DBLP:conf/acl/PoriaCHMZM17} employ an LSTM to capture context-level features from surrounding utterances.
\citet{DBLP:conf/naacl/HazarikaPZCMZ18,DBLP:conf/emnlp/HazarikaPMCZ18,DBLP:conf/aaai/JiaoLK20} use memory networks to capture contextual features.
\citet{DBLP:conf/aaai/MajumderPHMGC19} use GRUs to capture speaker, context and emotion features.
\citet{DBLP:conf/emnlp/JiaoLK20} introduce a conversation completion task based on unsupervised data to benefit the ERC task.
\citet{DBLP:conf/emnlp/GhosalMGMP20,DBLP:conf/ijcai/ZhaoZL22} utilize GRUs to fuse commonsense knowledge and capture complex interactions in the dialogue.
\citet{DBLP:conf/acl/HuWH20} 
propose a cognitive-inspired network that uses multi-turn reasoning modules to capture implicit emotional clues in conversations.
In this paper, we propose a supervised adversarial contrastive learning framework with contextual adversarial training to learn class-spread structured representations for better emotion recognition.

\subsubsection{Graph-based Methods}
Graph-based methods 
\cite{DBLP:conf/emnlp/GhosalMPCG19,DBLP:conf/ijcai/ZhangWSLZZ19,DBLP:conf/emnlp/IshiwatariYMG20,DBLP:conf/acl/ShenWYQ20,DBLP:conf/acl/HuLZJ20,DBLP:conf/icassp/HuHWJM22,DBLP:conf/ijcai/BaoMWZH22}
usually design a specific graph structure to capture complex dependencies in the conversation.
For example, \citet{DBLP:conf/emnlp/GhosalMPCG19,DBLP:conf/ijcai/ZhangWSLZZ19,DBLP:conf/acl/ShenWYQ20}  leverage GNNs to capture complex interactions in a conversation.
In order to simultaneously consider speaker interactions and sequence information, \citet{DBLP:conf/emnlp/IshiwatariYMG20} introduce a positional encoding module into RGAT.
\citet{DBLP:conf/acl/HuLZJ20,DBLP:conf/icassp/HuHWJM22} respectively design a graph-based fusion method that can simultaneously fuse multimodal knowledge and contextual features.

\subsubsection{Transformer-based Methods}
Transformer-based methods 
\citep{DBLP:conf/emnlp/ZhongWM19,DBLP:conf/sigdial/WangZMWX20,DBLP:conf/aaai/ShenCQX21,DBLP:conf/acl/ZhuP0ZH20,DBLP:conf/emnlp/Li00W21,DBLP:conf/emnlp/LeeC21,DBLP:conf/naacl/LeeL22,DBLP:conf/aaai/LiYQ22,DBLP:conf/emnlp/SongXH22}  usually exploit general knowledge in pre-trained language models \citep{DBLP:conf/naacl/DevlinCLT19,DBLP:journals/corr/abs-1907-11692,DBLP:conf/emnlp/0001HDZJMS22}, and model the conversation by a Transformer-based architecture.
For example,
\citet{DBLP:conf/emnlp/ZhongWM19} 
design a Transformer with graph attention to incorporate commonsense knowledge and contextual features.
\citet{DBLP:conf/sigdial/WangZMWX20} use a Transformer with an LSTM-CRF module to learn emotion consistency.
\citet{DBLP:conf/aaai/ShenCQX21} adopt a modified XLNet to deal with longer context and multi-party structures.
\citet{DBLP:conf/emnlp/LeeC21} leverage LSTM and GCN to enhance BERT's ability of context modeling.
\citet{DBLP:conf/aaai/YangSMC22} apply curriculum learning to deal with the learning problem of difficult samples.
\citet{DBLP:conf/aaai/LiYQ22}
utilize a supervised contrastive term and a response generation task to enhance BART's ability for ERC. 

\begin{table*}[t]
\centering  
\resizebox{\linewidth}{!}{$
\begin{tabular}{l|ccccccc|c|ccccccc|c}
  \hline
  \multicolumn{1}{c|}{\multirow{2}{*}{\makecell[c]{Optimization \\ Objectives}}} & \multicolumn{8}{c|}{{Attack Strength} $\epsilon$ (IEMOCAP)} 
  &\multicolumn{8}{c}{{Attack Strength} $\epsilon$ (MELD)}  \\
  \cline{2-17} 
  & 0 & 0.125 & 0.25 & 0.5 & 1 & 2 & 4 & Avg. ($\epsilon>0$)
  & 0 & 1 & 2 & 4 & 8 & 16 & 32 & Avg. ($\epsilon>0$)    \\
\hline
CE & 68.17 & 63.50 & 60.23 & 55.20 & 46.01 & 35.49 & 27.83 & 48.04
& 65.64 & 54.78 & 47.59 & 37.66 & 28.71 & 22.29 & 18.54 & 34.93   \\
CE with AT & 68.55 & 63.10 & 59.98 & 55.00 & 48.76 & 39.95 & 32.31 & 49.85
& 65.69 & 54.29 & 48.00 & 40.59 & 33.68 & 28.80 & 26.09 & 38.58  \\
\textbf{SACL} & \textbf{69.22} & \textbf{64.59} & \textbf{61.76} & \textbf{57.41} & \textbf{49.81} & \textbf{40.77} & \textbf{32.80} & \textbf{51.19}
& \textbf{66.45} & \textbf{62.03} & \textbf{59.96} & \textbf{56.35} & \textbf{50.88} & \textbf{44.69} & \textbf{37.69} & \textbf{51.93}  \\
Improve & +0.67 & +1.09 & +1.53 & +2.21 & +1.05 & +0.82 & +0.49 & +1.34
& +0.76 & +7.25 & +11.96 & +15.76 & +17.20 & +15.89 & +11.60 & +13.35  \\
\hline
  \end{tabular}
  $}
  \caption{
  Context robustness results against different optimization objectives on IEMOCAP and MELD. We report the robust weighted-F1 scores under different attack strengths.
  }
  \label{tab:appen:robusts1}
\end{table*}
\begin{figure*}[t]
  \centering
    \includegraphics[width=0.9\linewidth]{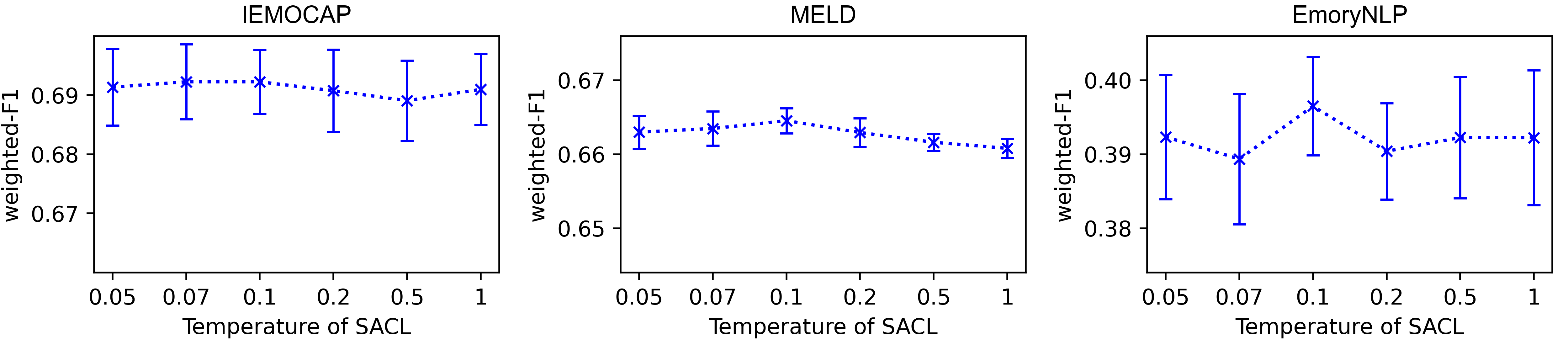}
    \caption{Classification performances of SACL-LSTM against different temperature coefficients on three datasets.}
  \label{fig:variant}
\end{figure*}

\subsection{Contrastive Learning and Adversarial Training}
\subsubsection{Contrastive Learning}
Contrastive learning is a representation learning technique to learn generalized embeddings such that similar data sample pairs are close while dissimilar sample pairs stay far apart \cite{DBLP:conf/cvpr/ChopraHL05}. 
\citet{DBLP:conf/nips/Sohn16,DBLP:journals/corr/abs-1807-03748,DBLP:conf/nips/BachmanHB19,DBLP:conf/eccv/TianKI20,DBLP:conf/icml/Henaff20,DBLP:conf/icml/ChenK0H20} utilize self-supervised contrastive learning to learn powerful representations. But these self-supervised techniques are generally limited by the risk of sampling bias and non-trivial data augmentation.
\citet{DBLP:conf/iclr/0001ZXH21} propose prototypical contrastive learning to encode the semantic structure of data into the embedding space.
\citet{DBLP:conf/nips/KimTH20,DBLP:conf/nips/JiangCCW20,DBLP:conf/nips/FanLCZG21} add instance-wise adversarial examples during self-supervised contrastive learning to improve model robustness.
Recently, \citet{khosla2020supervised,gunel2020supervised} use supervised contrastive learning to avoid the above risks and boost performance on downstream tasks by introducing label-level supervised signals. 
\citet{DBLP:conf/cvpr/00230W0W21,lopez2022supervised} use supervised contrastive learning over prototype-label embeddings to learn representations for classification.
\citet{DBLP:conf/naacl/LinMCYCC22} employ supervised contrastive learning and CE-based adversarial training to learn domain-adaptive features for low-resource rumor detection.
In this paper, we propose a supervised adversarial contrastive learning framework with contextual adversarial training to learn class-spread structured representations for classification on context-dependent data.

\subsubsection{Adversarial Training}
Adversarial training is a widely used regularization 
method to improve model robustness by generating adversarial examples with a min-max training recipe \cite{DBLP:journals/corr/SzegedyZSBEGF13}.
For example,
\citet{DBLP:journals/corr/SzegedyZSBEGF13} train neural networks on a mixture of adversarial examples and clean examples.
\citet{DBLP:journals/corr/GoodfellowSS14} further propose a fast gradient sign method to produce adversarial examples during training.
\citet{DBLP:conf/iclr/MiyatoDG17} extend adversarial and virtual adversarial training to the text domain by applying perturbations to the word embeddings.
After that, there are many variants established for supervised/semi-supervised learning \cite{DBLP:conf/nips/ShafahiNG0DSDTG19,DBLP:conf/nips/ZhangZLZ019,DBLP:conf/nips/QinMGKDFDSK19,DBLP:conf/acl/JiangHCLGZ20,DBLP:conf/iclr/ZhuCGSGL20}.

\section{Experimental Setups}  \label{sec:appendix:setups}
We report the detailed hyperparameter settings of SACL-LSTM on three datasets in Table~\ref{tab:appendix:param}.
The class weights in the CE loss are applied to alleviate the class imbalance issue and are set by their relative ratios in the train and validation sets, except for MELD, which presents a poor effect.
For MELD and EmoryNLP, we use focal loss \citep{DBLP:conf/iccv/LinGGHD17}, a modified version of the CE loss, to balance the weights of easy and hard samples during training.

\begin{table}[t]
\centering
\resizebox{\linewidth}{!}{$
\begin{tabular}{l|c|c|c}
\hline 
\multicolumn{1}{c|}{Hyperparameter}    &  IEMOCAP & MELD & EmoryNLP  \\ \hline
Embedding size \textbf{$d_u$} & 1024 & 1024 & 1024 \\ 
Hidden size $d_h$ & 128 & 128 & 128 \\ 
Number of LSTM layers & 2 & 1 & 1 \\
Threshold $\xi$ of Dual-LSTM &  2 & 2 & 2 \\
Perturbation radius $\epsilon$ of CAT &  5 & 5 & 0.5 \\ 
Perturbation rate of CAT & 1 & 1 & 0.1  \\
Norm constraint $L_q$ of CAT  & $L_2$ & $L_2$ & $L_2$ \\
Trade-off weight $\lambda$ of SACL & 0.05 & 0.1 & 0.5 \\ 
Trade-off weight $\lambda^{\text{r-adv}}$ of SACL & 0.5 & 0.05 & 0.1 \\ 
Temperature $\tau$ and $\tau^{\text{r-adv}}$ of SACL & 0.1 & 0.1 & 0.1 \\
\hline 
Number of epochs & 100 & 100 & 100 \\
Patience & 20 & 20 & 20 \\ 
Mini-batch size & 2 & 16 & 16 \\
Gradient accumulation steps & 16 & 2 & 2\\
Learning rate   & $1e^{-4}$ & $1e^{-4}$  & $5e^{-5}$   \\ 
Weight decay & $2e^{-4}$  & $2e^{-4}$ & $2e^{-4}$ \\ 
Dropout & 0.2 & 0.2 & 0.2 \\
Maximum token length & 200 & 200 & 200 \\
Flag of class weight & True & False & True \\ 
Factor of sample weight & - & 1 & 1 \\
\hline
\end{tabular}
$}
\caption{Hyperparameter settings of SACL-LSTM
on three datasets.
}
\label{tab:appendix:param}
\end{table} 

\section{Experimental Results}
\subsection{Results of Context Robustness Evaluation} \label{sec:app:robust}
The detailed results of context robustness evaluation on IEMOCAP and MELD are listed in Table~\ref{tab:appen:robusts1}.

\subsection{Parameter Analysis}
\label{sec:variant}
Figure~\ref{fig:variant} illustrates the effect of the temperature parameter in SACL framework on the ERC task.

\end{document}